\documentclass[10pt,twocolumn,letterpaper]{article}

\usepackage[margin=0.75in,top=1in]{geometry}
\usepackage{times}
\usepackage{epsfig}
\usepackage{graphicx}
\usepackage{amsmath}
\usepackage{amssymb}
\usepackage[dvipsnames]{xcolor}
\usepackage{multirow}
\usepackage{subcaption} 
\usepackage{cuted}
\usepackage{capt-of}
\usepackage{fancyhdr,graphicx}
\usepackage{graphicx}     
\usepackage{caption}      
\usepackage{rotating}     
\usepackage{multicol}     
\usepackage{pgffor}       
\usepackage{adjustbox}    
\usepackage{subcaption}   
\usepackage{bm}
\usepackage{tikz}

\definecolor{linkblue}{rgb}{0.21,0.49,0.74}
\usepackage[pagebackref,breaklinks,colorlinks,allcolors=linkblue]{hyperref}

\title{Auto-Regressive Transformation for Image Alignment}

\author{
Kanggeon Lee$^{1}$, Soochahn Lee$^{2}$\thanks{Corresponding authors}, Kyoung Mu Lee$^{1}$\footnotemark[1] \\
$^{1}$ASRI, Dept. of ECE, Seoul National University, Korea \\
$^{2}$Dept. of Electronics Engineering, Kookmin University, Korea 
\\
\footnotesize\texttt{dlrkdrjs97@snu.ac.kr, sclee@kookmin.ac.kr, kyoungmu@snu.ac.kr}
}

\date{}

\begin{document}

\maketitle

\begin{abstract}
 Existing methods for image alignment struggle in cases involving feature-sparse regions, extreme scale and field-of-view differences, and large deformations, often resulting in suboptimal accuracy.
Robustness to these challenges can be improved through iterative refinement of the transform field while focusing on critical regions in multi-scale image representations.
We thus propose Auto-Regressive Transformation (\textsc{ART}), a novel method that iteratively estimates the coarse-to-fine transformations through an auto-regressive pipeline. 
Leveraging hierarchical multi-scale features, our network refines the transform field parameters using randomly sampled points at each scale.
By incorporating guidance from the cross-attention layer, the model focuses on critical regions, ensuring accurate alignment even in challenging, feature-limited conditions.
Extensive experiments demonstrate that \textsc{ART} significantly outperforms state-of-the-art methods on planar images and achieves comparable performance on 3D scene images, establishing it as a powerful and versatile solution for precise image alignment.
\end{abstract}


\begin{figure}[t]
    \centering
    \begin{subfigure}[b]{0.29\columnwidth}
        \centering
        \includegraphics[height=4.75cm]{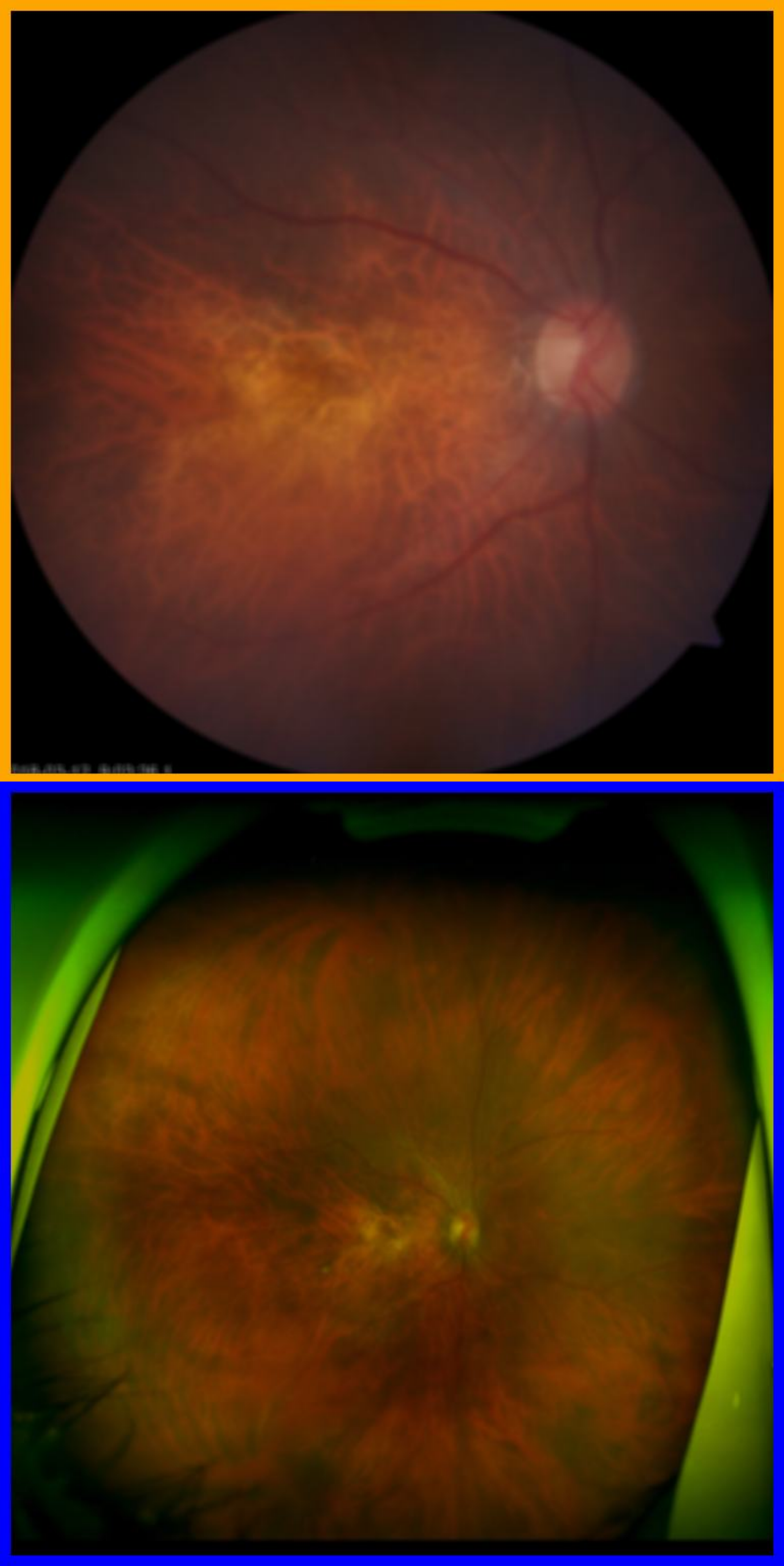}
        \caption*{\footnotesize Image Pair}
    \end{subfigure}%
    \begin{subfigure}[b]{0.35\columnwidth}
        \centering
        \includegraphics[height=4.75cm]{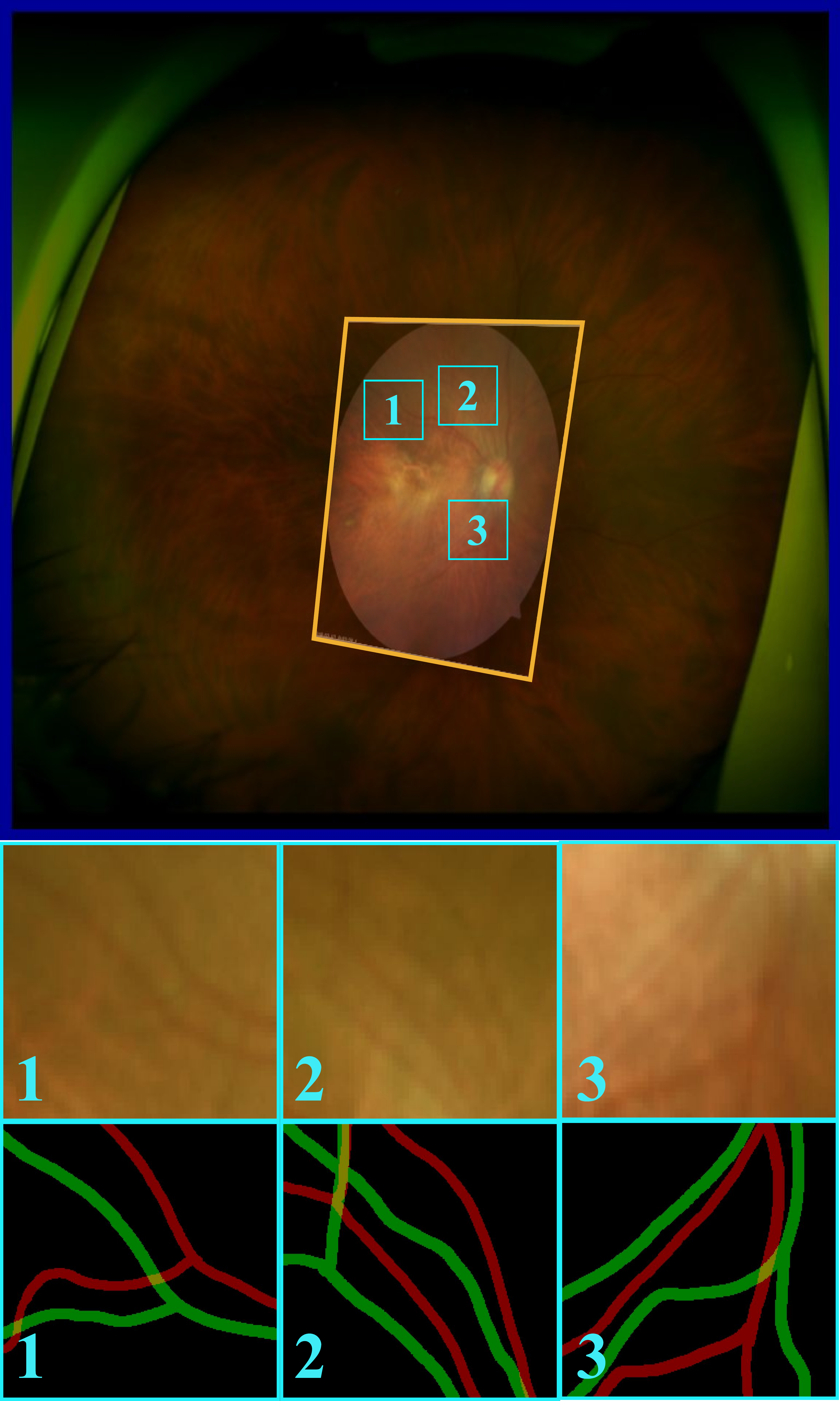}
        \caption*{\footnotesize GeoFormer~\cite{liu2023geometrized}}
    \end{subfigure}%
    \begin{subfigure}[b]{0.35\columnwidth}
        \centering
        \includegraphics[height=4.75cm]{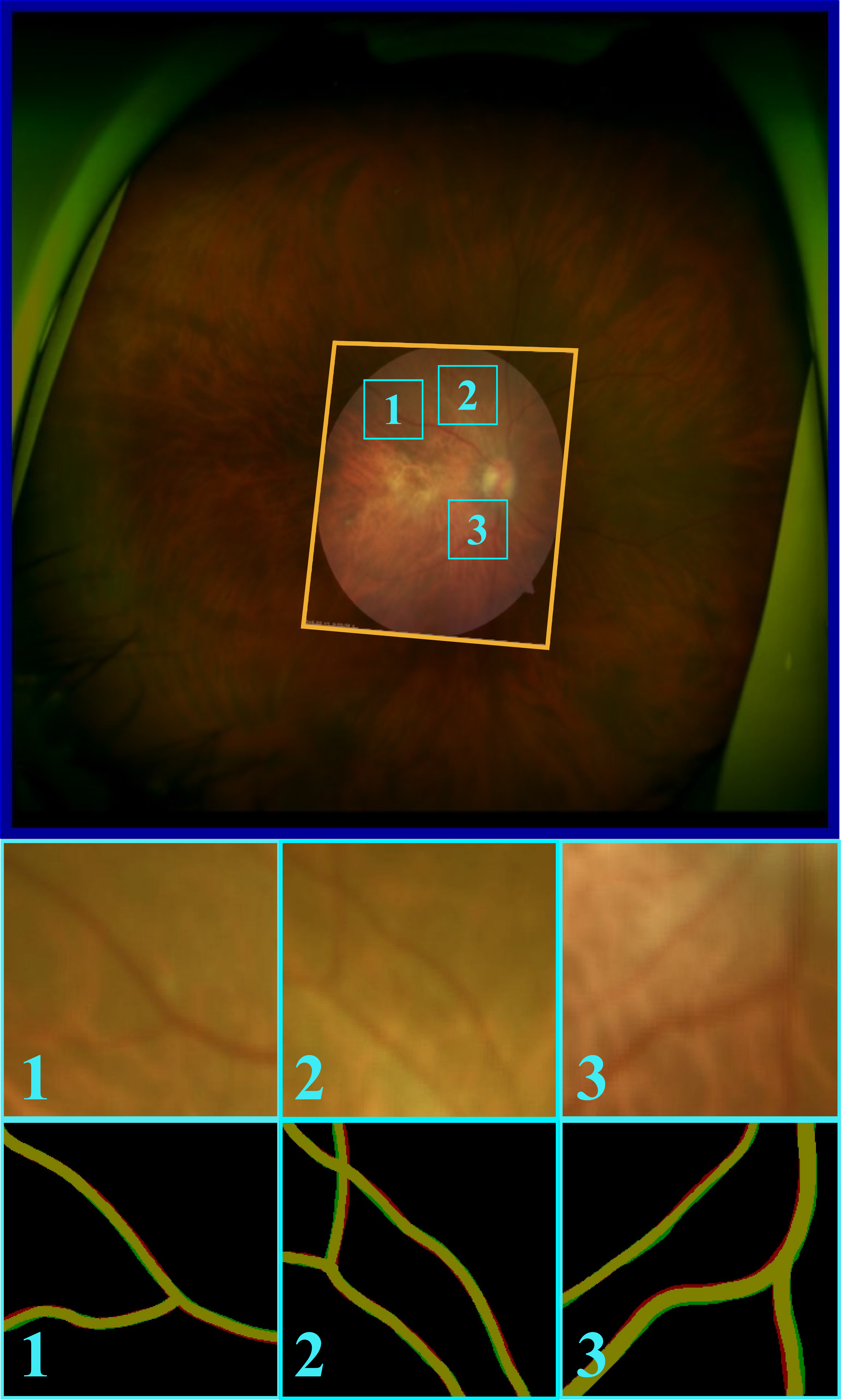}
        \caption*{\footnotesize \textsc{ART} (ours)}
    \end{subfigure}

    \caption{
        \textbf{Alignment Results in Challenging Scenarios.}
        For image pairs with sparse features, scale differences, deformations, degradations, and domain shifts, our method performs coarse-to-fine auto-regressive transformation refinement, achieving accurate alignment even in challenging scenarios where state-of-the-art methods struggle.
        The zoomed-in boxes show the local alignment results, and the highlighted vessel image below illustrates the intersection (yellow) between the two images (red and green).
    }
    \label{fig:FIG_TEASER}
\end{figure}
\vspace{-10pt}
\section{Introduction}
\label{sec:intro}

\begin{figure*}
    \centering
    \includegraphics[width=1.0\textwidth]{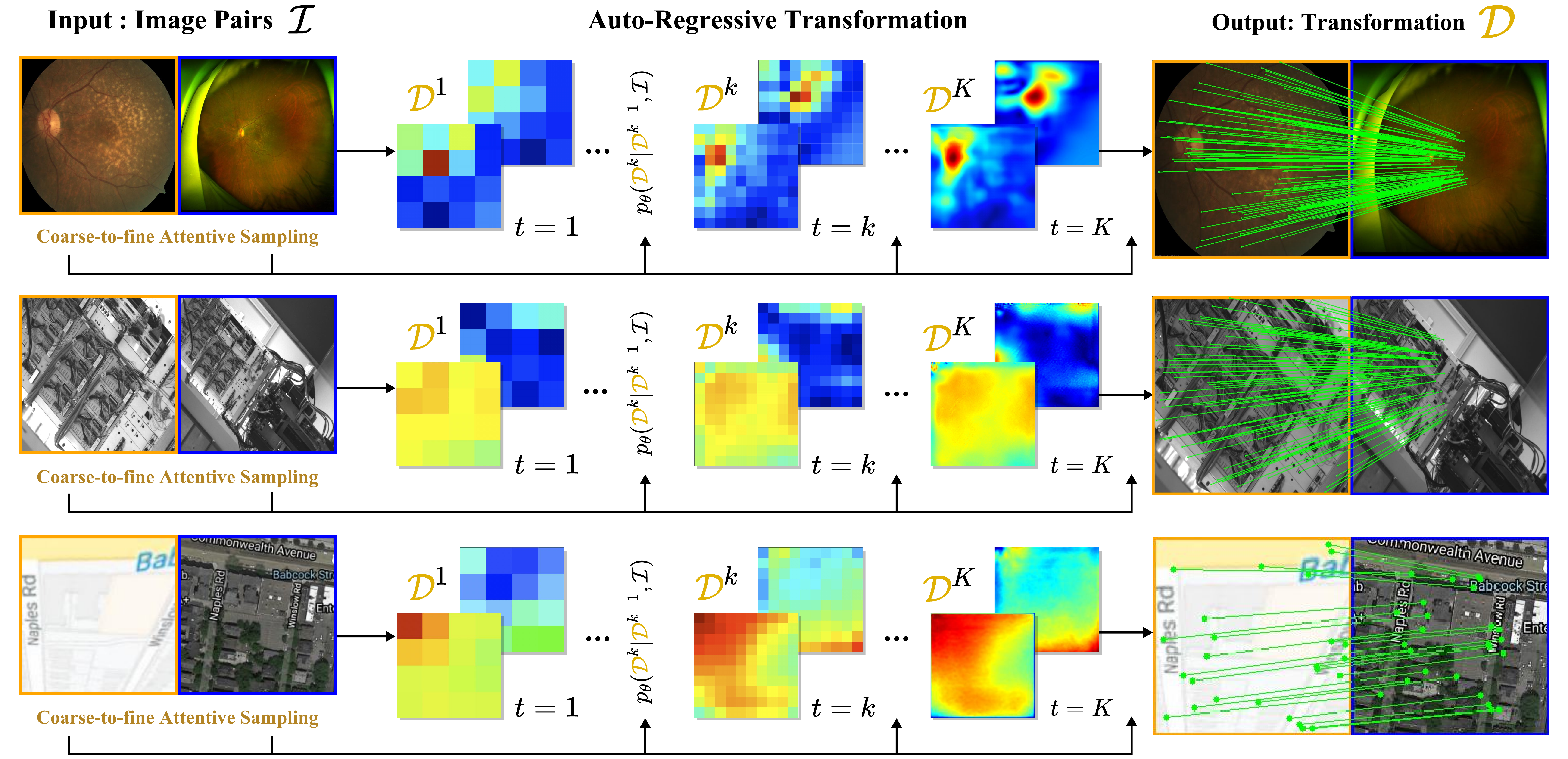}
    \caption{
    \textbf{Method Overview.} Auto-Regressive Transformation (\textsc{ART}) iteratively refines the transformation $\mathcal{D}$ for image pairs $\mathcal{I}$ in a coarse-to-fine manner. Its sampling strategy enables effective operation across diverse domains and datasets.
    } 
    \label{fig:FIG_OVERVIEW}
\end{figure*}

Image alignment is a fundamental problem in computer vision that involves registering images captured from different perspectives, times, or modalities. 
The process is essential for achieving seamless integration and analysis of images.
However, scale variations, structural deformations, and indistinct features complicate accurate alignment, requiring robust and adaptive methods.



Existing methods for image alignment often fail when 
(1) feature-based methods~\cite{hernandez2020rempe,sun2021loftr,liu2022semi,liu2023geometrized,Wang2024superjunction} struggle to detect keypoints due to homogeneous textures, low contrast, or weak features; 
(2) intensity-based methods~\cite{balakrishnan2019voxelmorph,kim2021cyclemorph,kim2022diffusemorph,chen2022transmorph,meng2023nicetransmorph,Ghahremani2024hvit} cannot handle large scale differences or deformations beyond their effective range; or 
(3) iterative refinement-based methods~\cite{besl1992method,cootes1995active,lee2019istn,de2019deep,zhao2021deep,zhu2024mcnet,meng2024correlationawarecoarsetofinemlpsdeformable,zhang2024adaptivecorrespondencescoringunsupervised,Ma2024IIPR} suffer from poor initialization, leading to slow convergence or suboptimal alignment results.

To reduce reliance on local feature matching, which primarily depends on tentative one-to-one correspondences, matching can be performed over larger appearance regions. 
This can be addressed by jointly estimating correspondences for sets of points.
To handle large scale differences, it is crucial to search across a wide range of scales.
This can be achieved by learning to infer transform field parameters within a coarse-to-fine framework, enabling the network to iteratively refine its estimates.
To improve robustness against poor initialization in iterative pipelines, non-parametric conditions should guide the refinement process. 
This can be done by incorporating global appearance cues from the input image pair into the sampling process.

Auto-Regressive Transformation (\textsc{ART}) is a novel image alignment framework robust to image pairs with large scale and field-of-view differences, deformations, and limited distinctive features, as shown in Fig.~\ref{fig:FIG_TEASER}.
\textsc{ART} employs an auto-regressive approach, iteratively sampling and refining local transform parameters by joint estimation for a set of points in a coarse-to-fine manner guided by multi-scale representations from a pyramid feature extraction network, as depicted in Fig.~\ref{fig:FIG_OVERVIEW}.
Moreover, by leveraging global appearance cues from the entire image pair as conditioning signals, \textsc{ART} achieves robustness to initialization.
Extensive evaluations demonstrate that \textsc{ART} significantly outperforms existing feature-based~\cite{detone2018superpoint,truong2019glampoints,rocco2020ncnet,sarlin2020superglue,hernandez2020rempe,sun2021loftr,liu2022semi,chen2022aspanformer,liu2023geometrized,sivaraman2024retinaregnetzeroshotapproachretinal}, intensity-based~\cite{de2019deep,cao2022iterative}, and iterative refinement-based methods~\cite{zhao2021deep,cao2022iterative,liu2022semi,zhu2024mcnet,sivaraman2024retinaregnetzeroshotapproachretinal} across various datasets.

Our contributions are as follows:
\begin{itemize}
    \item Coarse-to-fine auto-regressive modeling enables \textsc{ART} to handle substantial transformations between images.
    \item \textsc{ART} demonstrates state-of-the-art performance for a wide range of datasets with limited features, scale difference, large deformation, and considerable domain shift.
    \item \textsc{ART} can adapt to different complexity requirements by controlling the number of inference iterations.
\end{itemize}


\section{Related Works}
\label{sec:related}

\begin{figure*}[th!]
    \centering
    \includegraphics[width=1.0\textwidth]{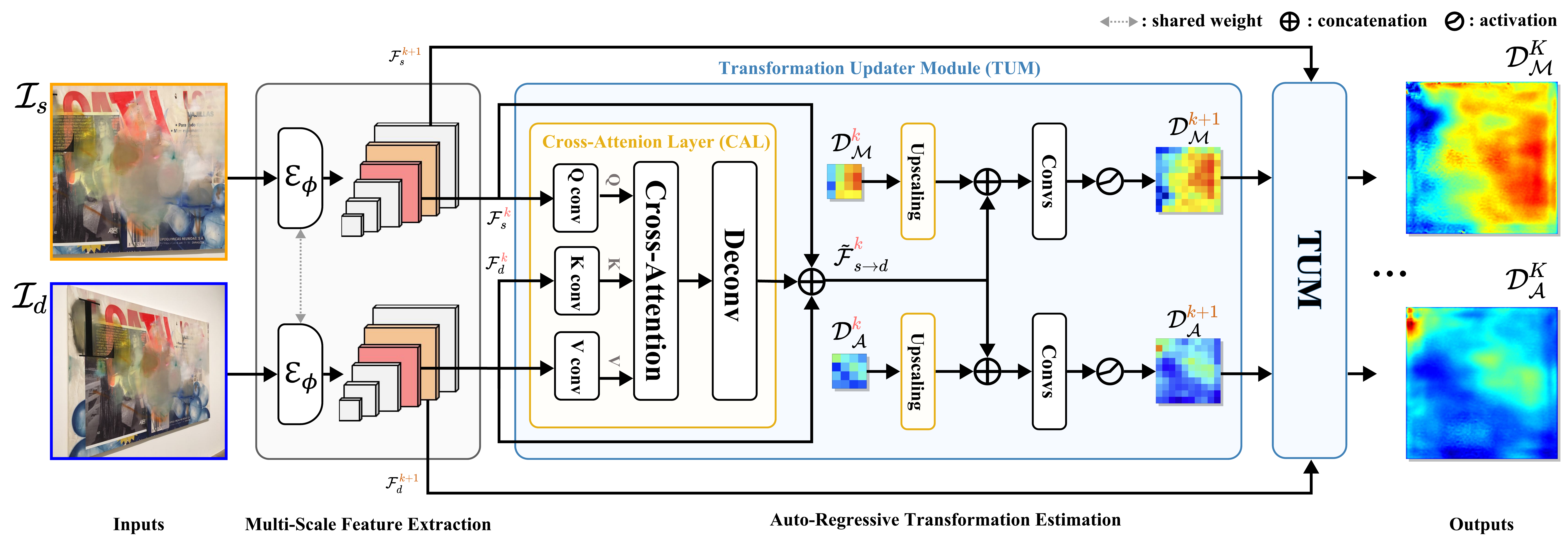}
    \caption{
    \textbf{Overall Framework.} \textsc{ART} first extracts multi-scale features $\mathcal{F}_s$ and $\mathcal{F}_d$ from the input image pair $\mathcal{I}_s$ and $\mathcal{I}_d$. At each sampling step $k$, the corresponding features, $\mathcal{F}_s^k$ and $\mathcal{F}_d^k$, are passed through the Cross-Attention Layer (CAL) to identify the correlated features that guide the network's focus on regions requiring refinement. The attentive feature map $\tilde{\mathcal{F}}_{s \rightarrow d}^k$ is then used to refine the transform field parameters $\mathcal{D}_{\mathcal{M}}^k$ and $\mathcal{D}_{\mathcal{A}}^k$ to $\mathcal{D}_{\mathcal{M}}^{k+1}$ and $\mathcal{D}_{\mathcal{A}}^{k+1}$ through multiple convolutional neural networks. This auto-regressive process continues until the initialized transform field parameters $\mathcal{D}_{\mathcal{M}}^0$ and $\mathcal{D}_{\mathcal{A}}^0$ reach the full resolution of the input image pair $\mathcal{I}_s$ and $\mathcal{I}_d$.
    } 
    \label{fig:FIG_NETWORK}
\end{figure*}


\vspace{-3pt}


\paragraph{Feature-based methods} align images by detecting and matching keypoints to estimate transformations. 
Traditional approaches~\cite{lowe2004distinctive,bay2006surf,rosten2008faster,calonder2010brief} are widely used for their robustness, while deep learning-based methods~\cite{detone2018superpoint,truong2019glampoints,revaud2019r2d2,liu2022semi} improve keypoint detection and description.

Further advancements enhance alignment performance.
SuperGlue~\cite{sarlin2020superglue} introduces graph neural networks for robust correspondence, while LightGlue~\cite{lindenberger2023lightglue} improves efficiency with a lightweight design.
However, these methods struggle in feature-sparse regions and under extreme distortions.

Detector-free methods estimate transformations directly from image pairs.
Deep homography estimation~\cite{detone2016deep} pioneered this approach, followed by NCNet~\cite{rocco2020ncnet}, which optimizes efficiency with sparse convolutions.
Optical flow-based models~\cite{xu2022gmflow,huang2022flowformer} estimate dense correspondences. 
Transformer-based methods such as LoFTR~\cite{sun2021loftr}, GeoFormer~\cite{liu2023geometrized}, and RoMa~\cite{edstedt2024roma}, as well as diffusion-based approaches like RetinaRegNet~\cite{sivaraman2024retinaregnetzeroshotapproachretinal} built on DIFT~\cite{tang2023emergent}, further enhance spatial reasoning.
However, these models often demand substantial computational resources and large-scale datasets for effective generalization.

\vspace{-12pt}
\paragraph{Intensity-based methods} align images by optimizing a transformation that minimizes pixel intensity differences using similarity metrics~\cite{viola1997alignment,zitova2003image}.
Traditional methods refine transform field parameters iteratively~\cite{maes1997multimodality,thirion1998image}.

Deep learning improves these approaches by directly predicting transformations, as seen in Deep Image Homography Estimation~\cite{detone2016deep} and Spatial Transformer Networks~\cite{jaderberg2015spatial}, while ISTN~\cite{lee2019istn} and REMPE~\cite{hernandez2020rempe} enhance flexibility and robustness.
These methods are widely applied in optical flow estimation~\cite{zhang2021separable, xu2022gmflow, huang2022flowformer} and medical image registration~\cite{cao2017deformable, hu2018weakly, xu2019deepatlas, balakrishnan2019voxelmorph, kim2021cyclemorph, Meng_2023}. 
However, they struggle with brightness variations, contrast differences, and modality changes, and can be computationally expensive for high-resolution images or complex transformations.

\vspace{-12pt}
\paragraph{Iterative refinement-based methods} progressively adjust transformations to improve alignment, inspired by traditional frameworks like RANSAC~\cite{fischler1981random} and ICP~\cite{besl1992method}. 
Early methods, such as Lucas-Kanade~\cite{lucas1981iterative}, employed gradient-based optimization but struggle with large deformations.
Deep learning models~\cite{zhao2021deep, cao2022iterative,zhu2024mcnet} refine alignment by employing multi-stage or recurrent processes. 
Diffusion-based approaches~\cite{wang2023posediffusion,zhang2024raydiffusion} further improve accuracy.

\vspace{-5pt}
\section{Proposed Method}
\label{sec:method}

\subsection{Problem Formulation}
Given a source image \( \mathcal{I}_s \) and a destination image \( \mathcal{I}_d \), both with spatial resolution \( (H, W) \), we denote their respective set of point coordinates as \(\mathcal{P}_s = \{(x_1, y_1), \dots, (x_n, y_n)\}\) and \(\mathcal{P}_d = \{(x_1', y_1'), \dots, (x_n', y_n')\}\), respectively. 

We can then define the locally linear point-wise transformations between \(\mathcal{P}_s\) and \(\mathcal{P}_d\) as:
\begin{equation}
\mathcal{P}_d = \mathcal{D}_{\mathcal{M}} \cdot \mathcal{P}_s + \mathcal{D}_{\mathcal{A}},
\end{equation}
where \( \mathcal{D}_{\mathcal{M}} \) and \( \mathcal{D}_{\mathcal{A}} \) represent the multiplicative and additive transform field parameters for point-wise scaling and translation, and the operations \(\cdot\) and \(+\) denote element-wise multiplication and addition, respectively.
Both \( \mathcal{D}_{\mathcal{M}} \) and \( \mathcal{D}_{\mathcal{A}} \) have shape \( (H, W, 2) \), where the last dimension corresponds to the \( x \) and \( y \)-axis components of the transformation.
That is, a point \((x_i, y_j)\) in \(\mathcal{P}_s\) is mapped to its corresponding point \((x'_i, y'_j)\) in \(\mathcal{P}_d\) as follows:
\begin{equation}\label{eq:displacement_path}
\begin{split}
    x_i' & = \mathcal{D}_{\mathcal{M}}[x_i,y_j,0] \times x_i + \mathcal{D}_{\mathcal{A}}[x_i,y_j,0], \\
    y_j' & = \mathcal{D}_{\mathcal{M}}[x_i,y_j,1] \times y_j + \mathcal{D}_{\mathcal{A}}[x_i,y_j,1].
\end{split}
\end{equation}
While each point transform is individually simple, the field as a whole can represent complex and flexible free-form deformations.



\subsection{Auto-Regressive Transformation}
To accurately estimate the transform field parameters \( \mathcal{D}_{\mathcal{M}} \) and \( \mathcal{D}_{\mathcal{A}} \) between \( \mathcal{I}_s \) and \( \mathcal{I}_d \), \textsc{ART} employs an auto-regressive coarse-to-fine refinement strategy, where transform field parameters are progressively updated through multiple steps, as was depicted in Fig.~\ref{fig:FIG_OVERVIEW}.

In the most coarse level, the \( (H_0,W_0,2) \) shaped transform field parameters \( \mathcal{D}_{\mathcal{M}}^0 \) and \( \mathcal{D}_{\mathcal{A}}^0 \) are initialized to \( \mathbf{1}_{H_0 \times W_0 \times 2}\) and \( \mathbf{0}_{H_0 \times W_0 \times 2}\), respectively.
Every iteration doubles the spatial resolution, so after \( k \) steps, \( \mathcal{D}_{\mathcal{M}}^k \) and \( \mathcal{D}_{\mathcal{A}}^k \) reach a spatial size \( 2^k \) times larger than \( \mathcal{D}_{\mathcal{M}}^0 \) and \( \mathcal{D}_{\mathcal{A}}^0 \), enabling the estimation of finer details.
This iterative refinement process enables the model to incrementally improve precise estimation at each sampling step \( k \) until reaching the final step \( K \), as follows:
\begin{equation}
(\mathcal{D}_{\mathcal{M}}^{k+1}, \mathcal{D}_{\mathcal{A}}^{k+1}) = \textsc{ART} (\mathcal{D}_{\mathcal{M}}^{k}, \mathcal{D}_{\mathcal{A}}^{k} | \mathcal{I}_s, \mathcal{I}_d).
\end{equation}


\begin{figure}[t] 
    \centering
    \includegraphics[width=1.0\columnwidth]{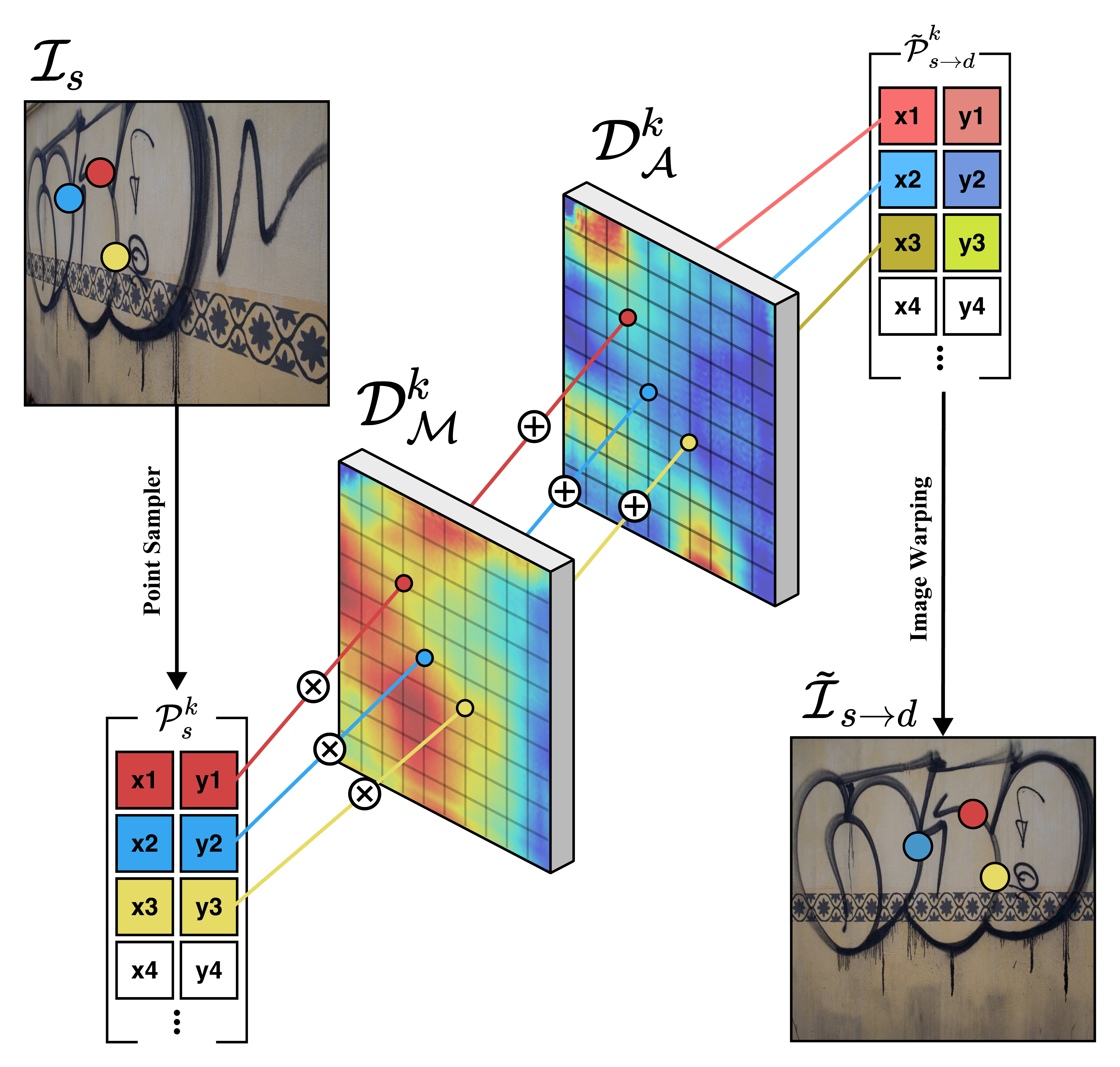}
    \caption{
    \textbf{Point-based Image Warping.} At sampling step \(k\), the extracted source points set \(\mathcal{P}_s^k\) is warped to \(\tilde{\mathcal{P}}_{s \rightarrow d}^k\) by sequentially multiplying with the corresponding values of the transform field parameter \(\mathcal{D}_{\mathcal{M}}^k\) and adding \(\mathcal{D}_{\mathcal{A}}^k\) for each point. These point pairs are then used to compute the warped image \(\tilde{\mathcal{I}}_{s \rightarrow d}\).
    }
    \label{fig:FIG_WARPING}
\end{figure}

\subsection{Architectural Details}
\textsc{ART} consists of two main components:  
(1) A multi-scale feature extractor that captures details from coarse to fine levels;
(2) A transformation updater module that autoregressively refines transform field parameters.
The entire network structure is depicted in Fig.~\ref{fig:FIG_NETWORK}.
Further details will be discussed in the following sections.

\vspace{-10pt}
\paragraph{Multi-Scale Feature Extractor}
The network \( \mathcal{E}_{\phi} \) as in \cite{Bai_2021,zhu2024mcnet} extracts multi-scale features with progressively increasing spatial resolutions from \( \mathcal{I}_s \) and \( \mathcal{I}_d \), as follows:
\begin{equation}\label{eq:displacement_path}
\begin{split}
    (\mathcal{F}_s^0,...,\mathcal{F}_s^k,...,\mathcal{F}_s^K)  & = \mathcal{E}_{\phi}(\mathcal{I}_s), \\
    (\mathcal{F}_d^0,...,\mathcal{F}_d^k,...,\mathcal{F}_d^K)  & = \mathcal{E}_{\phi}(\mathcal{I}_d), \\
\end{split}
\end{equation}
where \( k \) is the current and \( K \) is the maximum transform field parameters sampling step.
Each output’s spatial resolution doubles from the previous one, while the number of channels remains fixed. 
At each scale, feature maps from the previous resolution are progressively integrated, ensuring effective multi-scale feature fusion. 
This process enables the network to capture both fine-grained details and broader contextual information across resolutions.

\vspace{-10pt}
\paragraph{Transformation Updater Module}
At sampling step \( k \), the Cross-Attention Layer (CAL) first extracts the corresponding attentive feature map \( \tilde{\mathcal{F}}_{s \rightarrow d}^k \) by using the source feature map \( \mathcal{F}_s^k \) as the query and the destination feature map \( \mathcal{F}_d^k \) as the key and value as follows:
\begin{equation}
\tilde{\mathcal{F}}_{s \rightarrow d}^k = \texttt{Concat}[\mathcal{F}_s^k, \mathcal{F}_d^k, \text{CAL}(\mathcal{F}_s^k, \mathcal{F}_d^k)],
\end{equation}
where \texttt{Concat} represents concatenation of tensors.
To ensure computational efficiency with respect to the spatial size of the feature map, CAL applies a downsampling convolution to the query, key, and value features, followed by an upsampling deconvolution at the final stage.

The extracted attentive feature map \( \tilde{\mathcal{F}}_{s \rightarrow d}^k \) guides the subsequent network in determining where to focus, enabling the iterative update of \( \mathcal{D}_{\mathcal{M}}^{k+1} \) and \( \mathcal{D}_{\mathcal{A}}^{k+1} \) based on \( \mathcal{D}_{\mathcal{M}}^k \) and \( \mathcal{D}_{\mathcal{A}}^k \), which were derived from the previous step \( k-1 \).
This process can be expressed as follows:
\begin{equation}
\begin{split}
    \mathcal{D}_{\mathcal{M}}^{k+1}  & = \texttt{Conv}(\texttt{Concat}[^{\times 2}\mathcal{D}_{\mathcal{M}}^k, \tilde{\mathcal{F}}_{s \rightarrow d}^k]), \\
    \mathcal{D}_{\mathcal{A}}^{k+1}  & = \texttt{Conv}(\texttt{Concat}[^{\times 2}\mathcal{D}_{\mathcal{A}}^k, \tilde{\mathcal{F}}_{s \rightarrow d}^k]), \\
\end{split}
\end{equation}
Here, \texttt{Conv} denotes multiple convolutional layers with Leaky ReLU, \texttt{Concat} denotes tensor concatenation, and $^{\times 2}$ indicates bilinear upscaling by a factor of 2.

This iterative process is repeated in a coarse-to-fine manner to obtain the final transform field parameters \( \mathcal{D}_{\mathcal{M}}^K \) and \( \mathcal{D}_{\mathcal{A}}^K \), having the same spatial resolution as the input image.



\subsection{Image Warping}

At any \( k \)-th sampling step during autoregressive estimation, a set of source points \( \mathcal{P}_s^k \) selected by a point sampler can be warped as \( \tilde{\mathcal{P}}_{s \rightarrow d}^k = \mathcal{D}_{\mathcal{M}}^k \cdot \mathcal{P}_s^k + \mathcal{D}_{\mathcal{A}}^k \), as depicted in Fig.~\ref{fig:FIG_WARPING}. 

We can utilize these sets of source points \(\mathcal{P}_s^k\) and the corresponding points \(\tilde{\mathcal{P}}_{s \rightarrow d}^k\) to model the transform function from the source image \(\mathcal{I}_s\) to the destination image \(\mathcal{I}_d\) to get warped image \(\tilde{\mathcal{I}}_{s \rightarrow d}\).
This can be represented either as a linear warp~\cite{hartley2003multiple} for global changes or a quadratic warp~\cite{kim2022diffusemorph} for both global and local deformations.

\subsection{Training ART}
The end-to-end training loss \(\mathcal{L}\) of \textsc{ART} is defined as:
\begin{equation}\label{eq_full}
\mathcal{L} = \mathcal{L}_{\text{P}} + \lambda_{\text{R}}\mathcal{L}_{\text{R}},
\end{equation}
where \(\mathcal{L}_\text{P}\) and \(\mathcal{L}_\text{R}\) are the pixel matching loss, and regularization loss, respectively. 
\(\lambda_{\text{R}}\) controls the relative importance of the regularization loss.

\vspace{-10pt}
\paragraph{\textbf{Stochastic Pixel Matching Loss}}
\(\mathcal{L}_\text{P}\) computes the difference of warped source points set \( \tilde{\mathcal{P}}_{s \rightarrow d}^k\) with ground-truth destination points set \( \mathcal{P}_{d}^k\) for all \(0 < k  \leq K\) as follows:
\begin{equation}\label{eq:pml}
\mathcal{L}_{\text{P}} = \mathbb{E}_{k} \left\| \tilde{\mathcal{P}}_{s \rightarrow d}^k - \mathcal{P}_{d}^k \right\|_{2}^{2}.
\end{equation}
Note that we can utilize the point sampler used for image warping to stochastically select the source points set \( \mathcal{P}_s^k \).
This stochastic sampling, instead of using regular grid points or conventional keypoint detection techniques~\cite{lowe2004distinctive}, plays a key role in enabling the network to learn to robustly estimate the transforms at any particular scale.

\vspace{-12pt}
\paragraph{\textbf{Regularization Loss}} 
To complement the pixel matching loss, we define a regularization term to ensure that all warped points converge to their appropriate positions using homography matrix \( \mathcal{H} \), shaping the distribution of estimated correspondence points rather than directly predicting ground-truth coordinates, for all \(0 < k  \leq K\) as follows:
\begin{equation}
    \mathcal{L}_{\text{R}} = \mathbb{E}_{k} \left\| \mathcal{H}^k - \mathcal{H}_{GT}^k \right\|_{1}^{1},   
\end{equation}
where \(\mathcal{H}^k\) is computed from differentiable RANSAC~\cite{Brachmann2019DSACstar} with inlier threshold $2$ and \(\mathcal{H}_{GT}^k\) is the ground-truth with sampling step \( k \).

\section{Experiments}
\label{sec:experiment}

\begin{table}[t]
\renewcommand*{\arraystretch}{1.6}
\caption {\textbf{Datasets for Evaluation.}}
\label{tab:dataset}
\resizebox{1.0\columnwidth}{!}{
\Huge
\begin{tabular}{ccllr}
\hline
\multicolumn{1}{l}{\textbf{Category}} & \multicolumn{1}{l}{\textbf{Type}} & \textbf{Dataset}     & \textbf{Image Content}                                      & \textbf{Training Type} \\ \hline
\multirow{3}{*}{\textbf{Retinal}}     & \multirow{3}{*}{HR}      & KBSMC      & Image pairs of SFI and UWFI                                            & FS                     \\
                                      &                          & FIRE~\cite{hernandezmatas2017fire}        & Image pairs of SFI and SFI                                            & SS                     \\
                                      &                          & FLORI21~\cite{Li2021Flori}     & Image pairs of UWFI and UWFI                                               & SS                     \\ \hline
\multirow{6}{*}{\textbf{Scene}}      & \multirow{3}{*}{HR}  & HPatches~\cite{balntas2017hpatches}    & Planar images under varying illumination and viewpoint              & SS    \\ 
                                      &   & MegaDepth-1500~\cite{MegaDepthLi18}    & Outdoor scenery images under different lighting and perspective conditions              & SS    \\ 
                                      &   & ScanNet-1500~\cite{dai2017scannet}    & Indoor scenery images with real-world viewpoint and lighting variations                & SS    \\

\cline{2-5} 
                                      & \multirow{3}{*}{LR}      & GoogleEarth~\cite{zhao2021deep} & Satellite images of the earth's surface                                    & FS                     \\
                                      &                          & GoogleMap~\cite{zhao2021deep}   & Navigation map with satellite images                                    & FS                     \\
                                      &                          & MSCOCO~\cite{lin2015microsoftcoco}      & Common images in natural context               & FS                     \\ \hline
\end{tabular}}
\scriptsize{
FS and SS denote fully-supervised and self-supervised, respectively.\\
HR and LR denote high-resolution and low-resolution, respectively.
}
\end{table}

\subsection{Datasets}\label{subsec:dataset}
Evaluation of \textsc{ART} is performed across retinal and scene categories, as described in Tab.~\ref{tab:dataset}.

For retinal images, we evaluate \textsc{ART} on three datasets, comprising standard fundus images (SFI) and ultra-wide fundus images (UWFI). 
For cross-domain alignment, we use a private dataset from the Kangbuk Samsung Medical Center (KBSMC) Ophthalmology Department, collected between 2017 and 2019\footnote{This study adhered to the tenets of the Declaration of Helsinki and was approved by the Institutional Review Boards (IRB) of Kangbuk Samsung Hospital (No. KBSMC 2019-08-031). The study is a retrospective review of medical records, and the data were fully anonymized prior to processing. The IRB waived the requirement for informed consent.}, consisting of 3,744 SFI-UWFI pairs with scale differences of approximately \( \times 1 \sim \times 4 \), where ground truth transformation was manually annotated.
Additionally, we utilize the public datasets FIRE~\cite{hernandezmatas2017fire} and FLORI21~\cite{Li2021Flori} for in-domain alignment of SFI-SFI and UWFI-UWFI pairs.

For scene categories, we evaluate \textsc{ART} on HPatches~\cite{balntas2017hpatches} (planar images), 
MegaDepth-1500~\cite{MegaDepthLi18} (outdoor images),
ScanNet-1500~\cite{dai2017scannet} (indoor images),
GoogleEarth~\cite{zhao2021deep} (satellite images), GoogleMap~\cite{zhao2021deep} (navigation maps), and MSCOCO~\cite{lin2015microsoftcoco} (common images). 

\vspace{-3pt}
\subsection{Implementation Details}\label{subsec:implementation}

Here, we define high-resolution (HR) and low-resolution (LR) images as \(768 \times 768\) and \(192 \times 192\), respectively, based on their spatial dimensions.
KBSMC, FIRE~\cite{hernandezmatas2017fire},  FLORI21~\cite{Li2021Flori},  HPatches~\cite{balntas2017hpatches},
MegaDepth-1500~\cite{MegaDepthLi18}, and
ScanNet-1500~\cite{dai2017scannet} are assigned as HR type, while GoogleEarth~\cite{zhao2021deep},  GoogleMap~\cite{zhao2021deep}, and MSCOCO~\cite{lin2015microsoftcoco} are LR, respectively.
The original images may be resized to meet these definitions.
The number of inference steps \(K\) is set to $6$ for HR and $4$ for LR images, respectively.
The point sampler selected $100$ points, randomly for training and via a feature detector~\cite{lowe2004distinctive} for consistent testing.

\vspace{-10pt}
\paragraph{Common Setup}\label{subsec:setup}
We used the AdamW~\cite{loshchilov2019decoupled} optimizer with a learning rate of \(0.001\), \(\beta_1 = 0.9\), \(\beta_2 = 0.999\), and \(\epsilon = 10^{-8}\) to train \textsc{ART}, applying weight decay every \(100\)K iterations with a decay rate of \(0.01\). 
The model was trained for more than \(1\)K epochs using an NVIDIA A100 GPU.
We set \(\lambda_{\text{R}}\) to \(0.5\). 
For the fully-supervised (FS) training strategy in Tab.~\ref{tab:dataset}, we apply data augmentation by introducing random rotations and random photometric distortions, including variations in illumination, contrast, blur, and noise. 
For the self-supervised (SS) training strategy in Tab.~\ref{tab:dataset}, we additionally apply random transformations to the training image, along with the aforementioned augmentation.
We normalize the points in \(\mathcal{P}_{s}\) to have values in the range \([-1, 1]\).
We set the spatial dimensions of the initialized transform field parameters \(\mathcal{D}_{\mathcal{M}}^0\) and \(\mathcal{D}_{\mathcal{A}}^0\) to \(H_0 = 12\) and \(W_0 = 12\).


\begin{figure*}[t]
    \centering

    \def\imgwidth{0.235\textwidth}
    \def\labelwidth{0.035\textwidth}
    \def\rowlabel#1{%
      \begin{minipage}[c][\imgheight][c]{\labelwidth}
        \centering
        \rotatebox{90}{\scriptsize #1}
      \end{minipage}%
    }

    \newlength{\imgheight}
    \settoheight{\imgheight}{\includegraphics[width=\imgwidth]{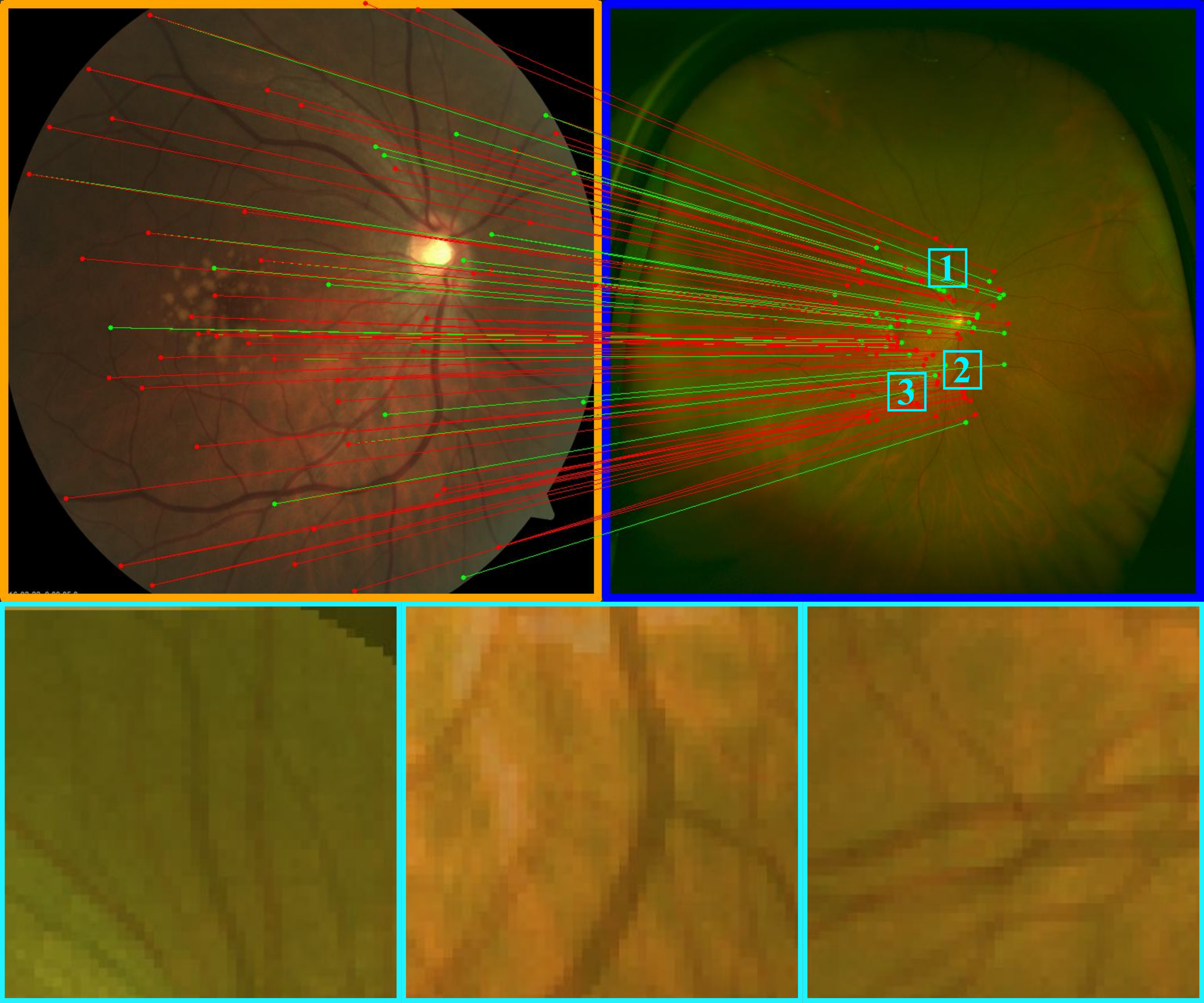}}

    \rowlabel{KBSMC}
    \begin{minipage}[c]{\imgwidth}\centering\includegraphics[width=\linewidth]{figs/fig_retinal_1_1.pdf}\end{minipage}%
    \begin{minipage}[c]{\imgwidth}\centering\includegraphics[width=\linewidth]{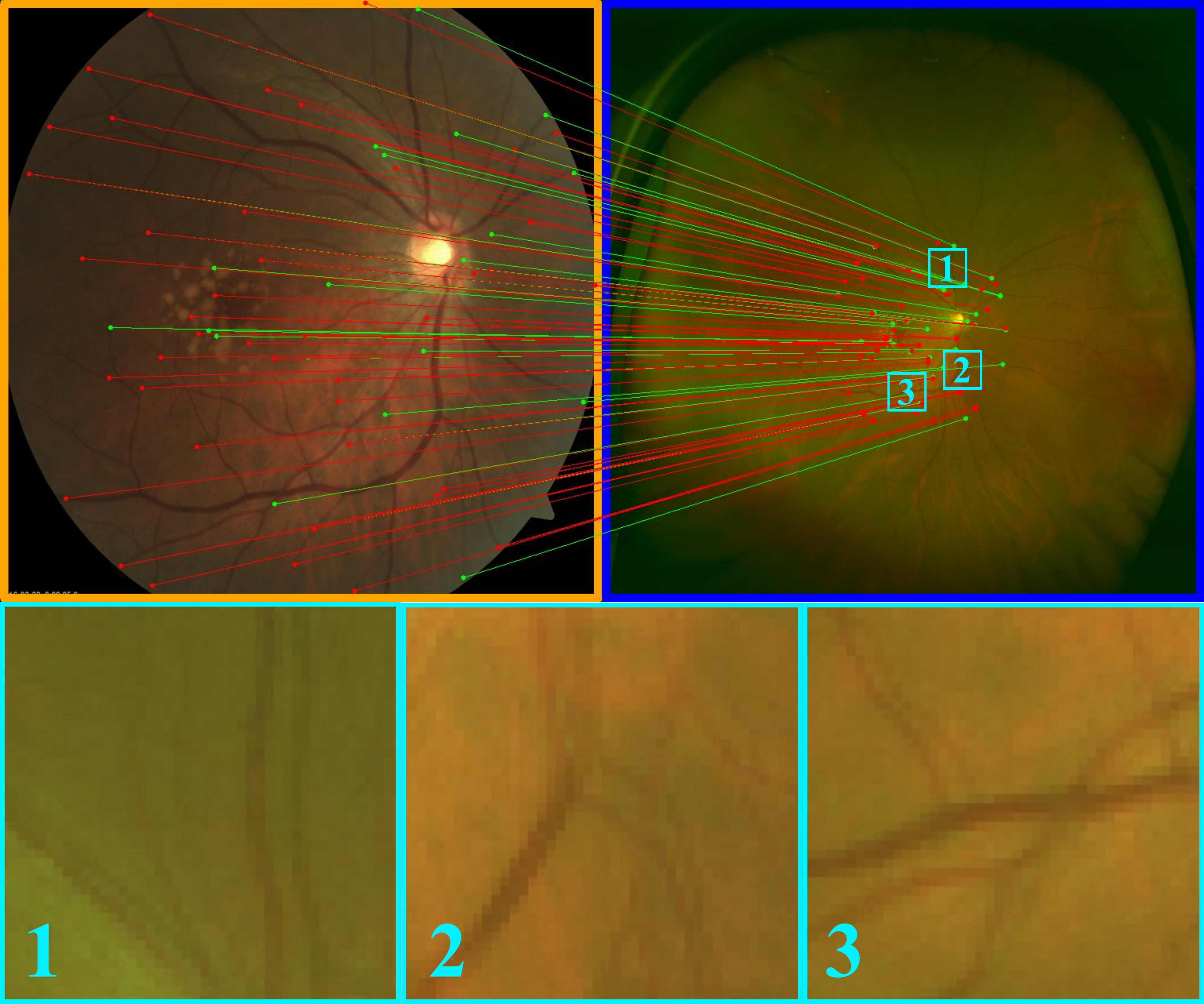}\end{minipage}%
    \begin{minipage}[c]{\imgwidth}\centering\includegraphics[width=\linewidth]{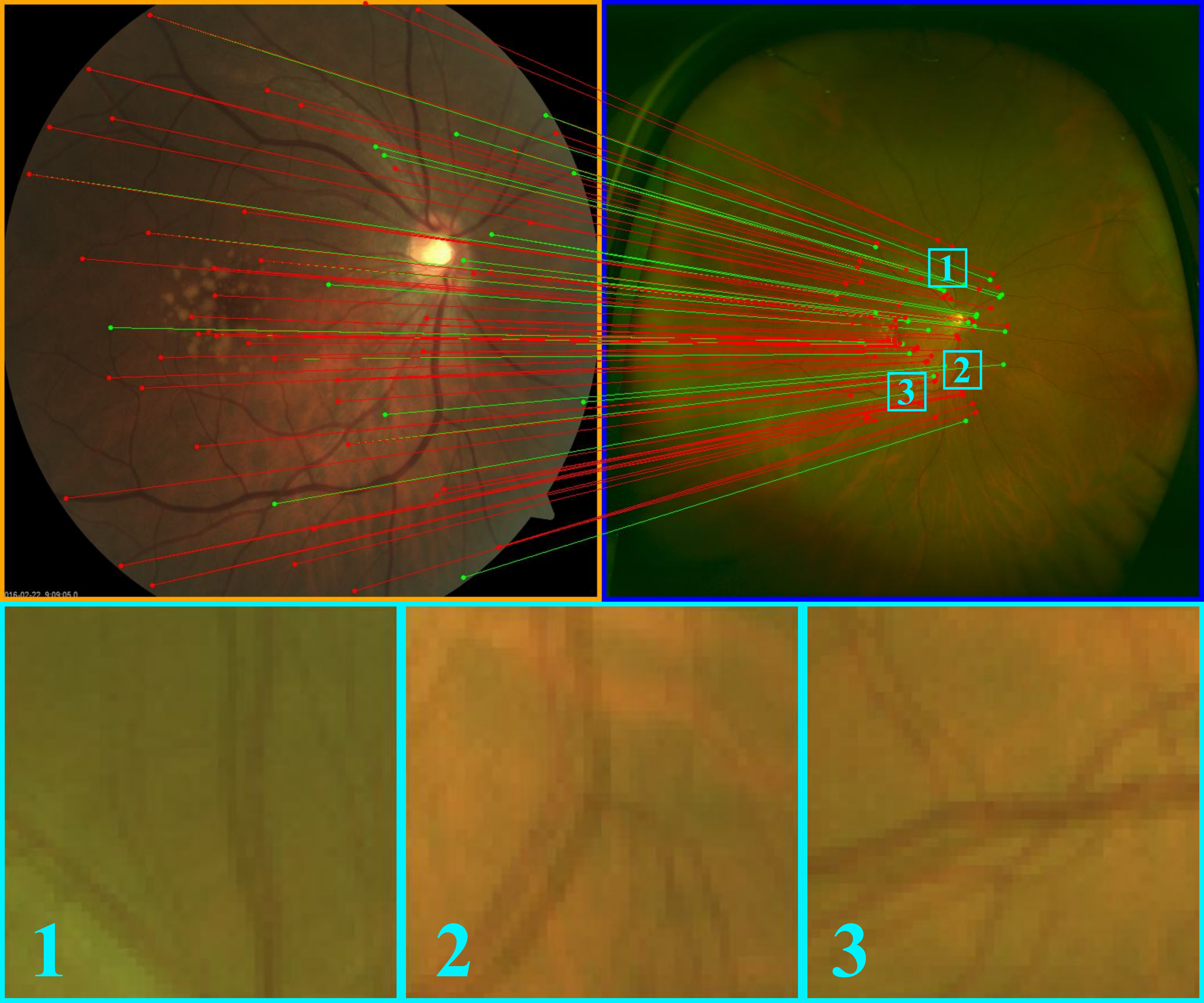}\end{minipage}%
    \begin{minipage}[c]{\imgwidth}\centering\includegraphics[width=\linewidth]{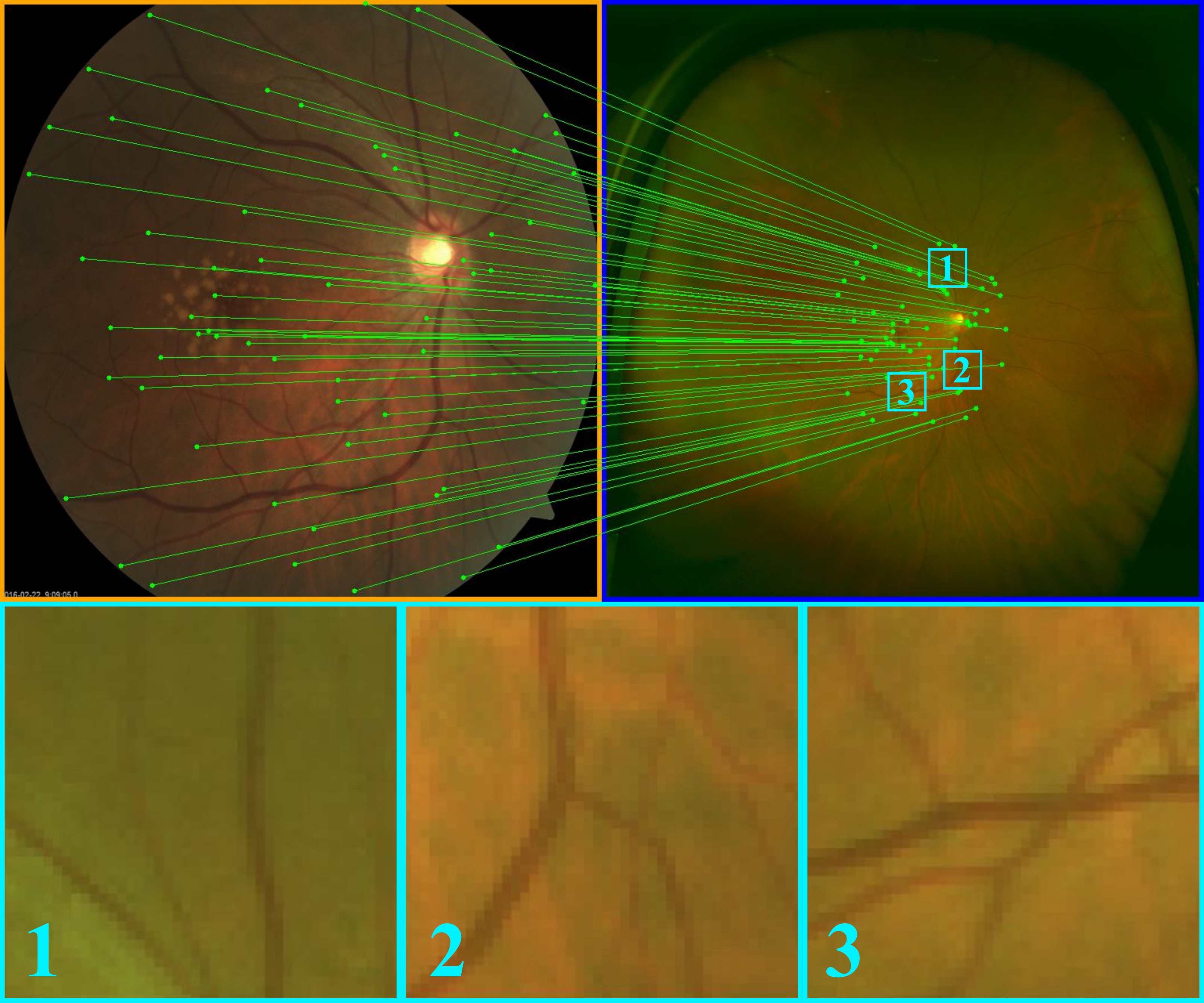}\end{minipage}%

    \rowlabel{FIRE~\cite{hernandezmatas2017fire}}
    \begin{minipage}[c]{\imgwidth}\centering\includegraphics[width=\linewidth]{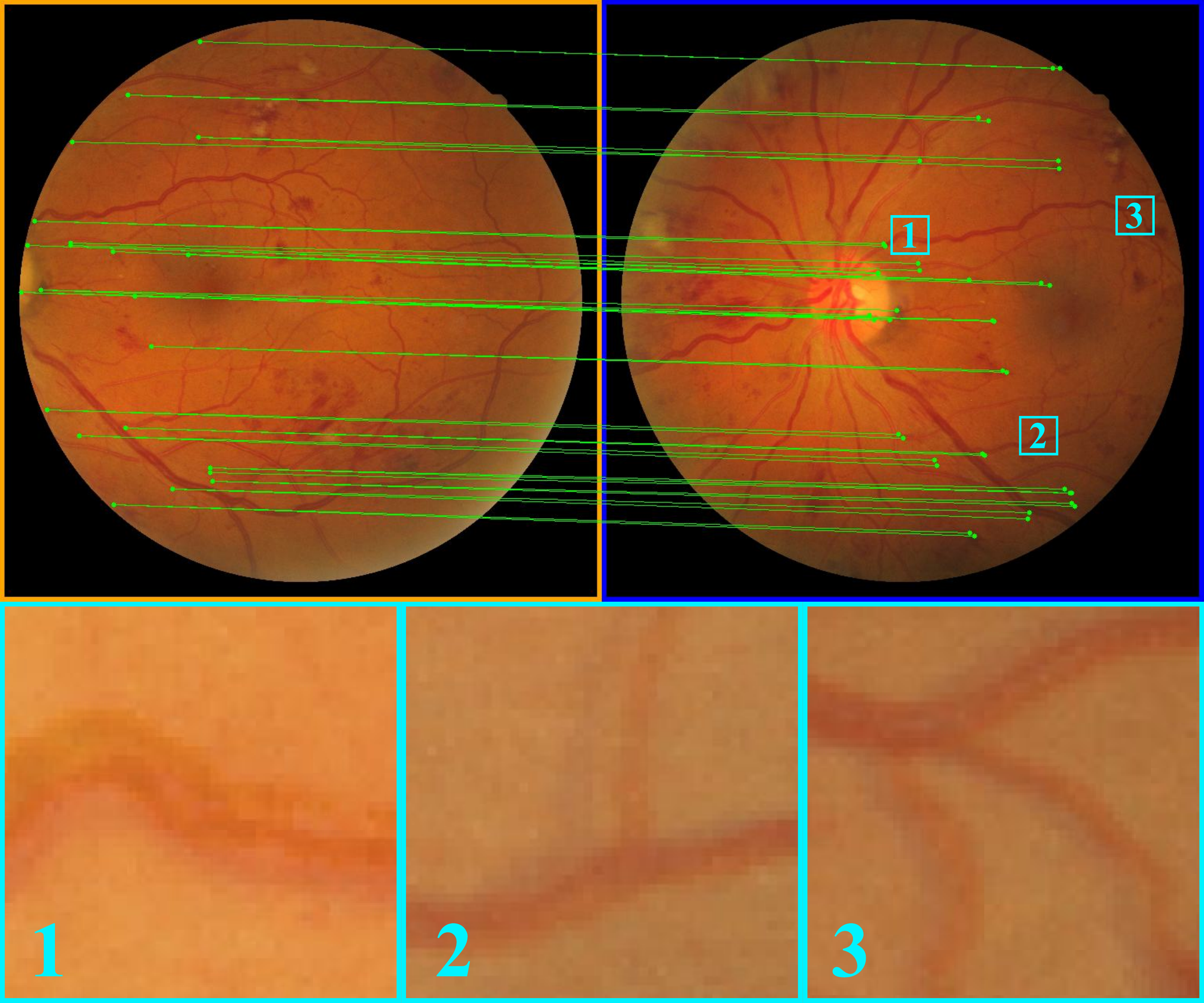}\end{minipage}%
    \begin{minipage}[c]{\imgwidth}\centering\includegraphics[width=\linewidth]{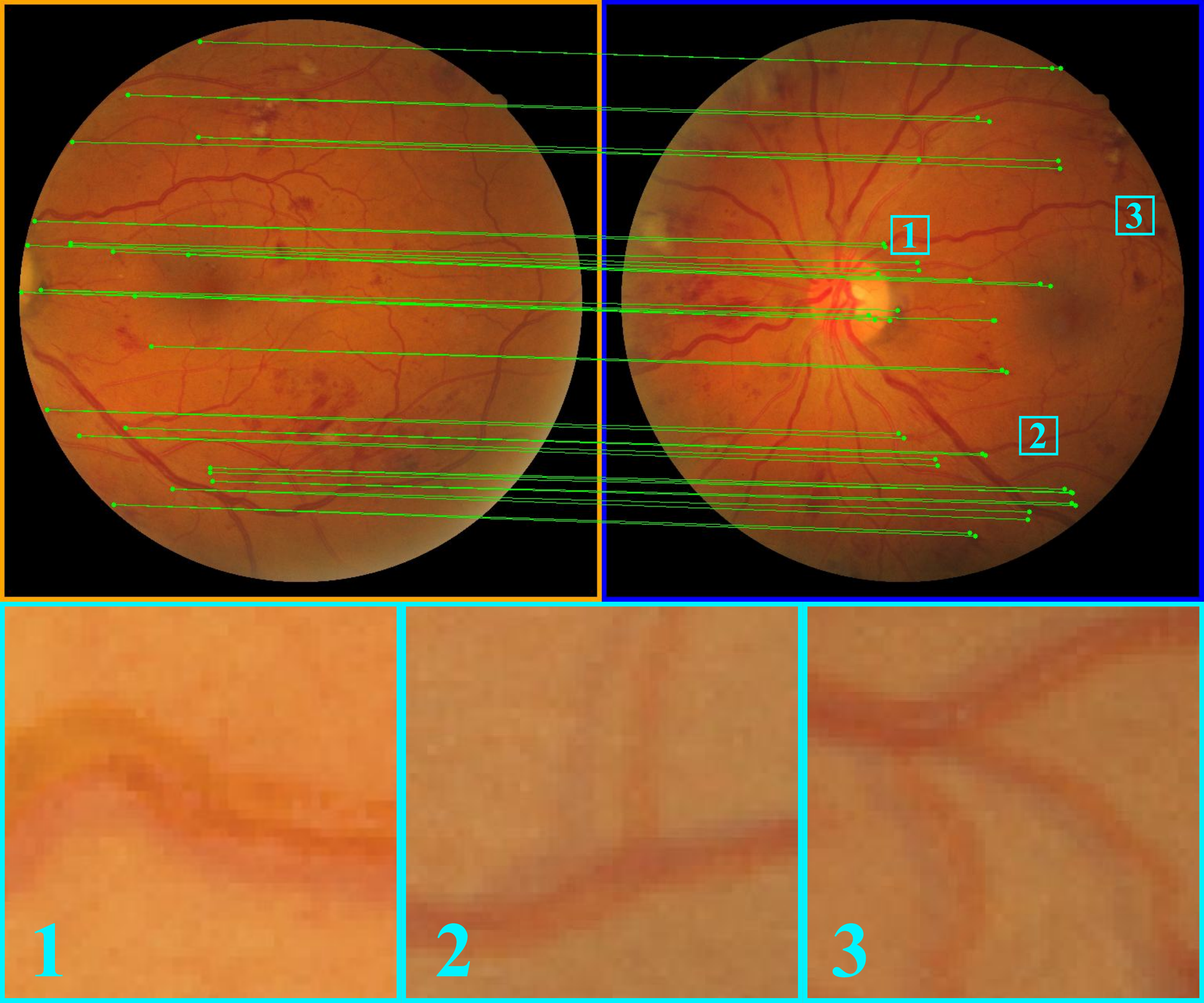}\end{minipage}%
    \begin{minipage}[c]{\imgwidth}\centering\includegraphics[width=\linewidth]{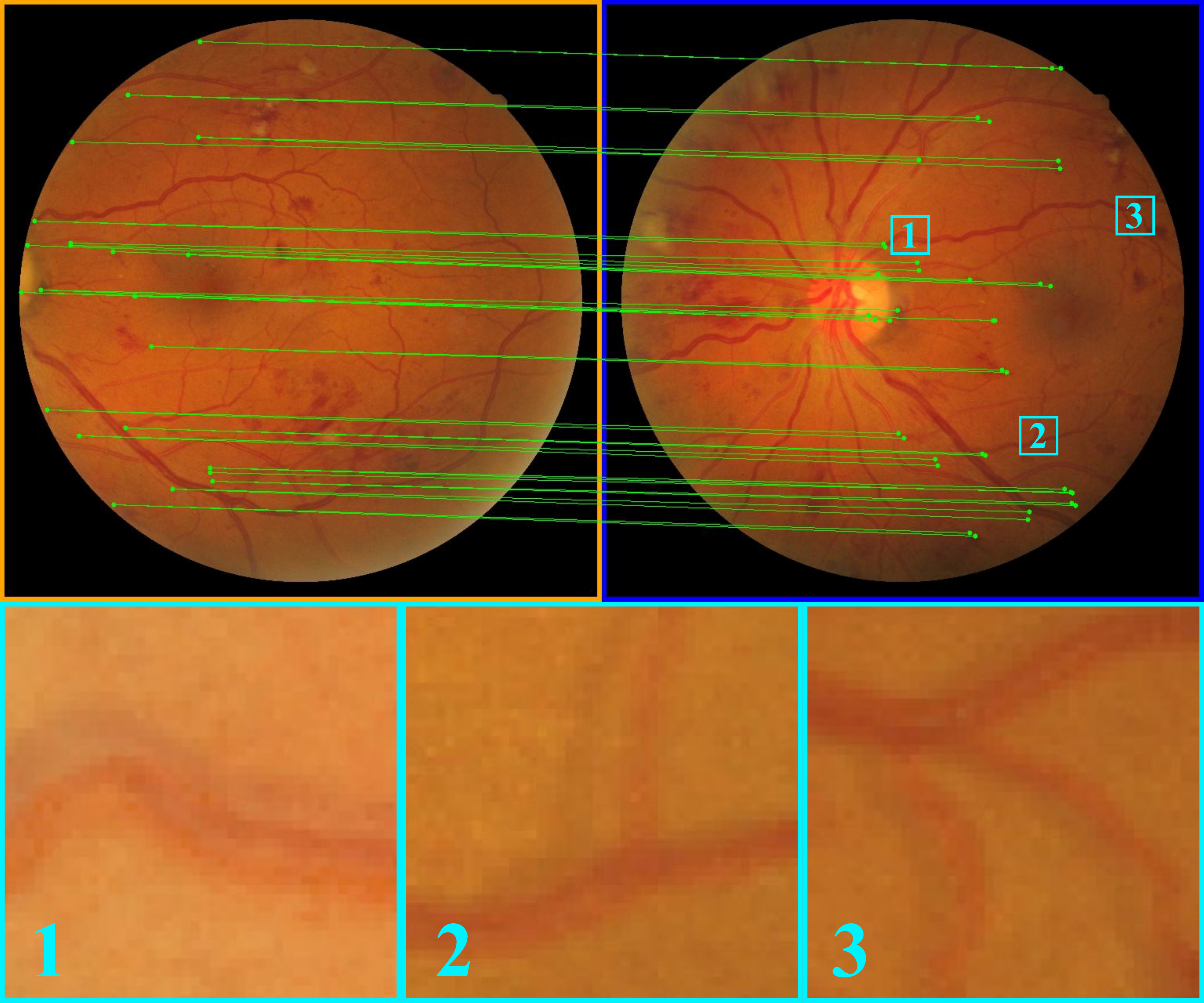}\end{minipage}%
    \begin{minipage}[c]{\imgwidth}\centering\includegraphics[width=\linewidth]{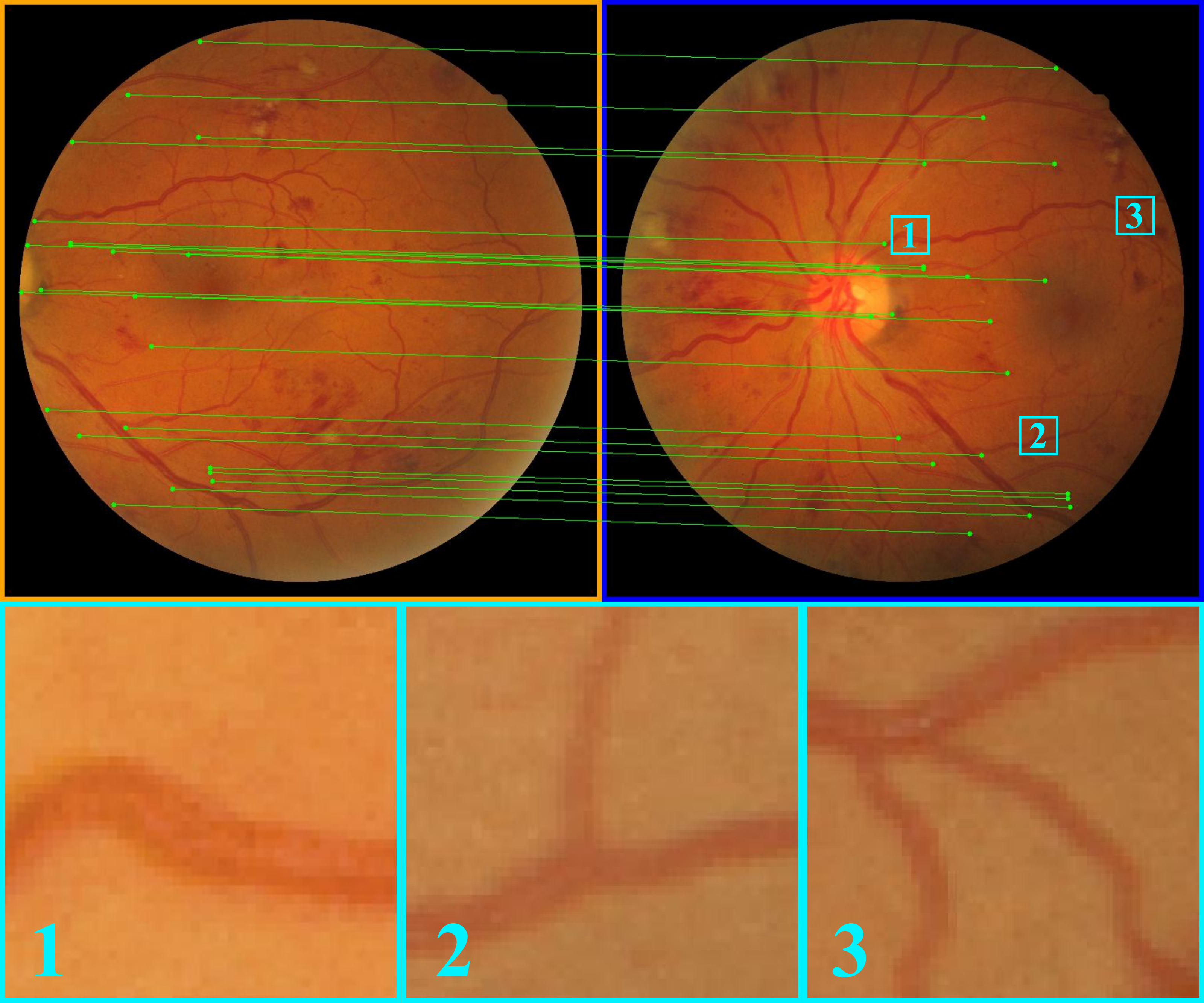}\end{minipage}%

    \rowlabel{FLORI21~\cite{Li2021Flori}}
    \begin{minipage}[c]{\imgwidth}\centering\includegraphics[width=\linewidth]{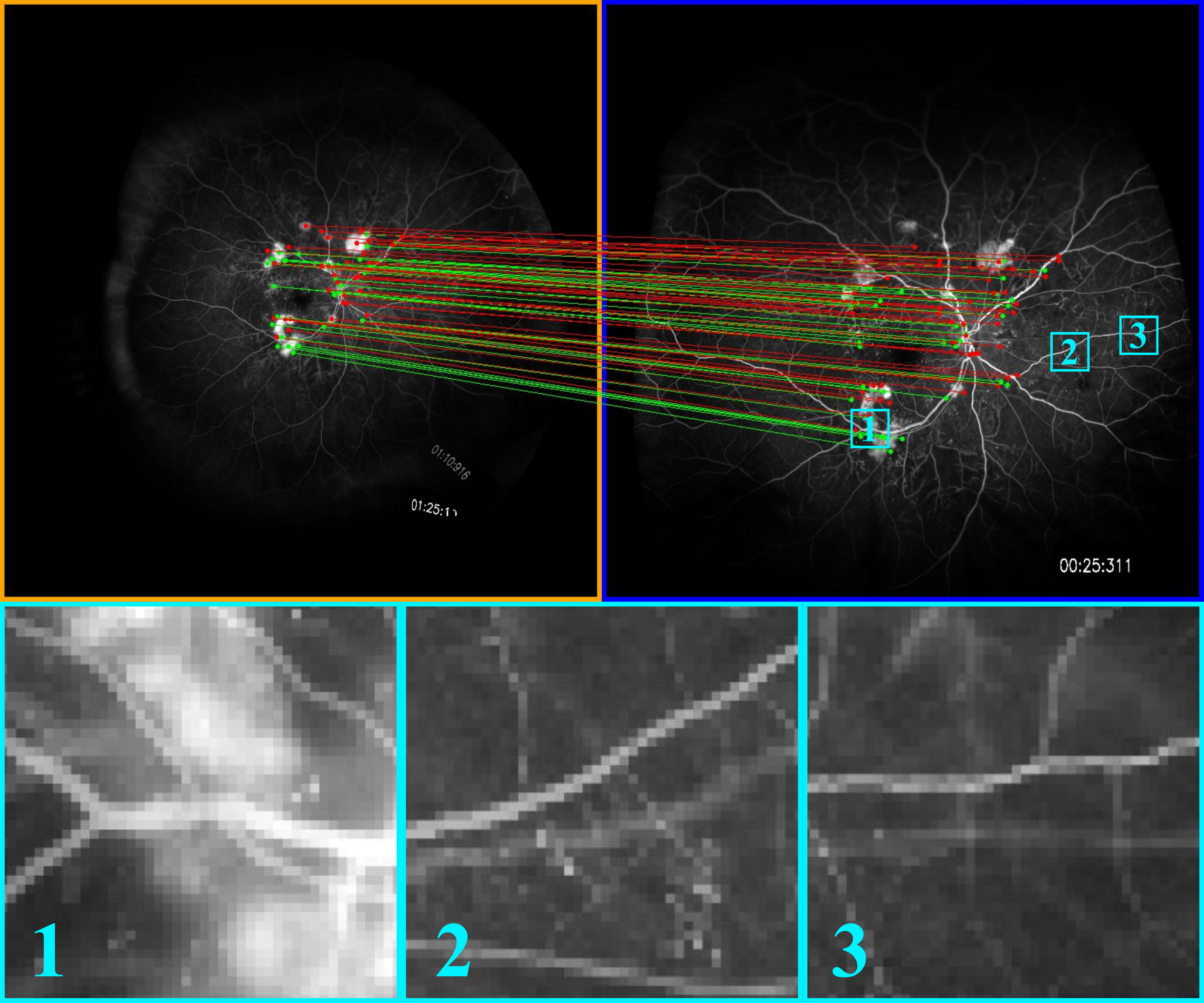}\end{minipage}%
    \begin{minipage}[c]{\imgwidth}\centering\includegraphics[width=\linewidth]{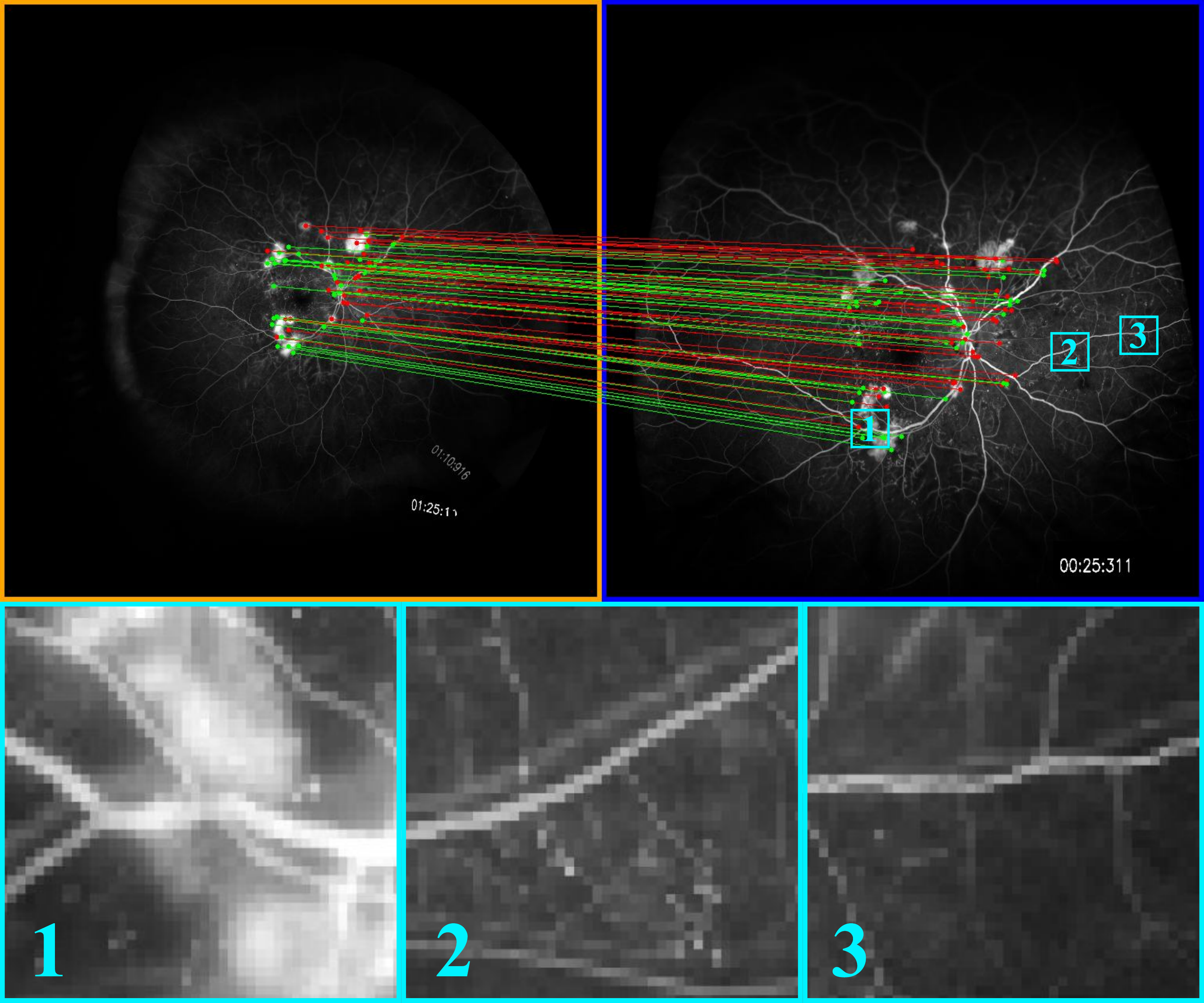}\end{minipage}%
    \begin{minipage}[c]{\imgwidth}\centering\includegraphics[width=\linewidth]{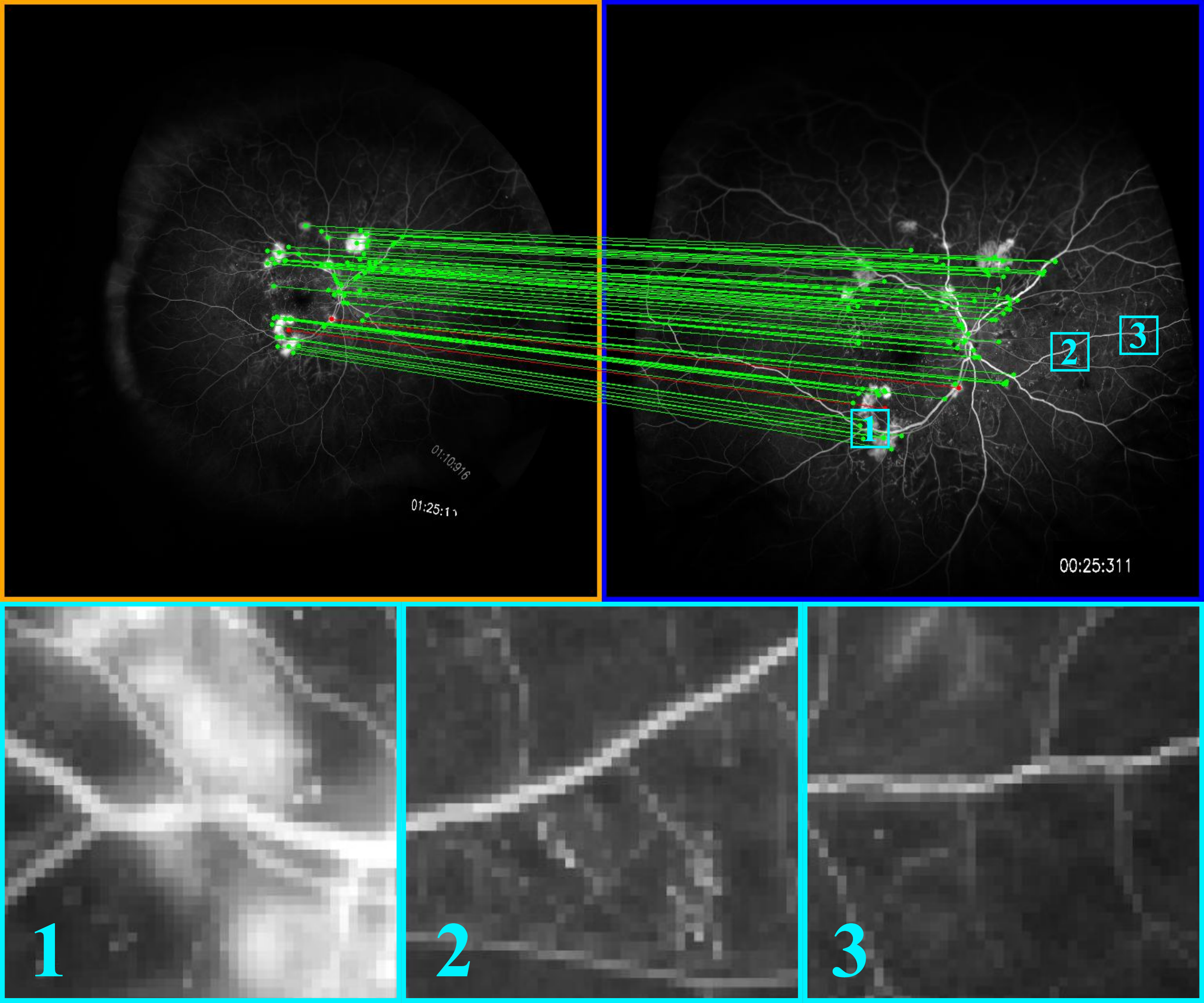}\end{minipage}%
    \begin{minipage}[c]{\imgwidth}\centering\includegraphics[width=\linewidth]{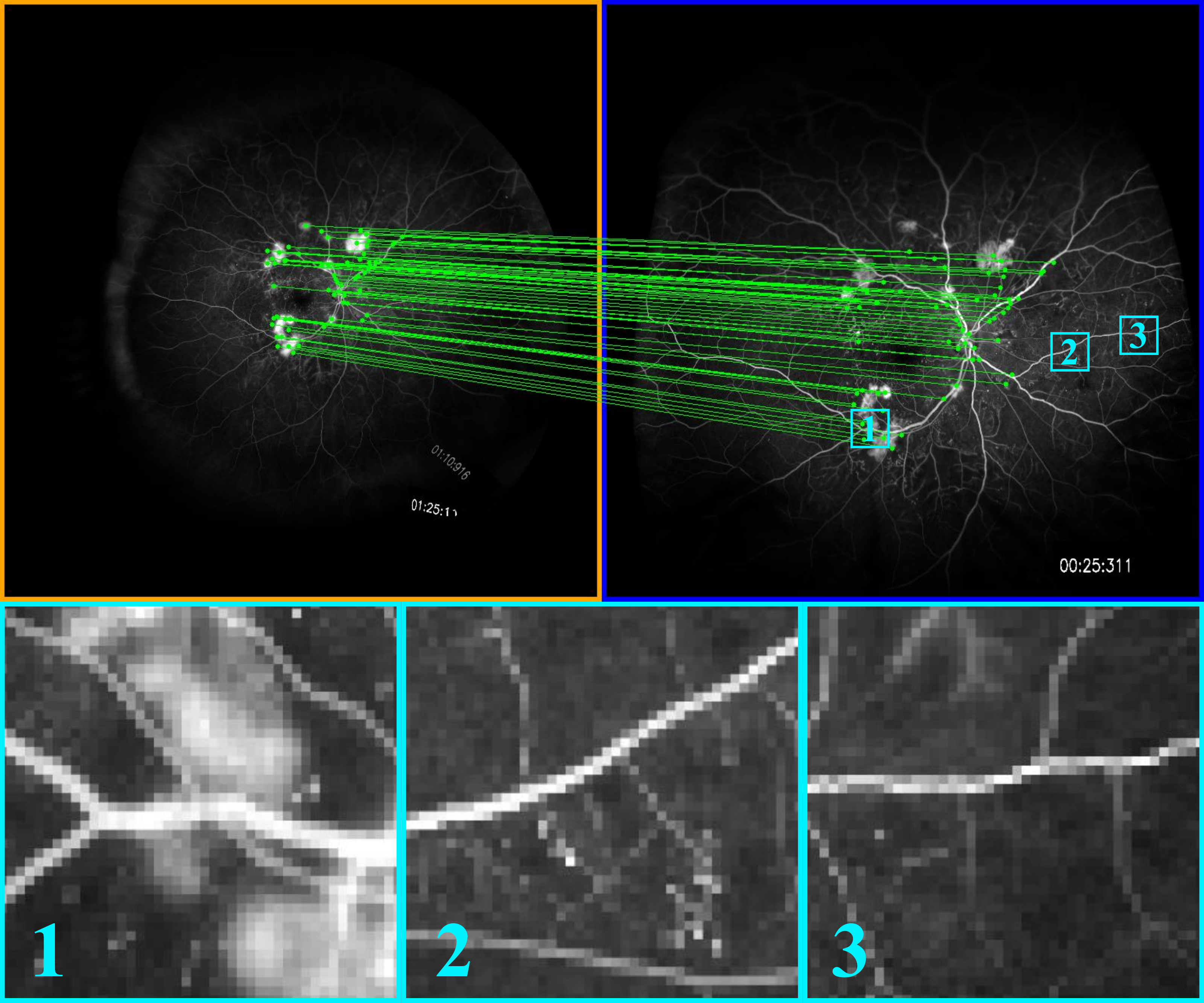}\end{minipage}%

    \vspace{0.4em}
    \begin{minipage}[b]{\labelwidth}
    \end{minipage}%
    \hspace*{0.64cm}
    \begin{minipage}[b]{\imgwidth}\centering\footnotesize SuperRetina~\cite{liu2022semi}\end{minipage}%
    \begin{minipage}[b]{\imgwidth}\centering\footnotesize GeoFormer~\cite{liu2023geometrized}\end{minipage}%
    \begin{minipage}[b]{\imgwidth}\centering\footnotesize RetinaRegNet~\cite{sivaraman2024retinaregnetzeroshotapproachretinal}\end{minipage}%
    \begin{minipage}[b]{\imgwidth}\centering\footnotesize \textsc{ART} (ours)\end{minipage}%

    \caption{
        \textbf{Qualitative Evaluation on Retinal Datasets.}
        Across various domains, \textsc{ART} robustly identifies sufficient matches compared to SuperRetina~\cite{liu2022semi}, GeoFormer~\cite{liu2023geometrized}, and RetinaRegNet~\cite{sivaraman2024retinaregnetzeroshotapproachretinal}.
        Correct and incorrect matches are shown in green and red, respectively. The zoomed-in boxes highlight overlaid local regions.
    }
    \label{fig:FIG_RETINAL}
\end{figure*}

\begin{table*}[]
\caption {\textbf{Quantitative Evaluation on Retinal Datasets.}}
\label{tab:retinal}
\renewcommand*{\arraystretch}{0.95}
\resizebox{1.0\textwidth}{!}{
\begin{tabular}{lcccccccc}
\hline
\multirow{2}{*}{\textbf{Methods}}                                                          
&   \multicolumn{2}{c}{\textbf{KBSMC}}                                          
& & \multicolumn{2}{c}{\textbf{FIRE}}                                             
& & \multicolumn{2}{c}{\textbf{FLORI21}}   \\ \cline{2-3} \cline{5-6} \cline{8-9} 
& \textit{Acceptable}\textsubscript{\scriptsize$\uparrow$}(\%)      & \textbf{mAUC}\textsubscript{\scriptsize$\uparrow$}                           & 
& \textit{Acceptable}\textsubscript{\scriptsize$\uparrow$}(\%)      & \textbf{mAUC}\textsubscript{\scriptsize$\uparrow$}                           &  
& \textit{Acceptable}\textsubscript{\scriptsize$\uparrow$}(\%)      & \textbf{mAUC}\textsubscript{\scriptsize$\uparrow$}                             \\ \hline
SuperPoint~\cite{detone2018superpoint}                               & 9.09                               & 8.7                                     &           & 94.78                                    & 67.3                               &  & 40                                 & 39.1                              \\
GLAMpoints~\cite{truong2019glampoints}                               & 9.89                               & 8.4                                     &           & 93.28                                    & 61.9                               &  & 33.33                              & 34.4                              \\
ISTN~\cite{lee2019istn}                                              & 20.86                              & 12.1                                    &           & 86.57                                    & 60.9                               &  & 53.33                              & 52.5                              \\
NCNet~\cite{rocco2020ncnet}                                          & 12.30                              & 9.6                                     &           & 86.57                                    & 61.4                               &  & 53.33                              & 50.8                              \\
SuperGlue~\cite{sarlin2020superglue}                                 & 24.06                              & 15.3                                    &           & 95.52                                    & 68.7                               &  & 80                                 & 59.8                              \\
REMPE~\cite{hernandez2020rempe}                                      & 22.46                              & 15.0                                    &           & 97.01                                    & 72.1                               &  & 73.33                              & 60.0                              \\
DLKFM~\cite{zhao2021deep}                                            & 22.73                              & 13.5                                    &           & 86.57                                    & 61.4                               &  & 40                                 & 40.1                              \\
LoFTR~\cite{sun2021loftr}                                            & 26.20                              & 16.9                                    &           & 97.01                                    & 71.5                               &  & 66.67                              & 51.5                              \\
IHN~\cite{cao2022iterative}                                          & 23.80                              & 14.5                                    &           & 88.81                                    & 63.5                               &  & 60                                 & 50.0                              \\
SuperRetina~\cite{liu2022semi}                                       & 34.76                              & 22.3                                    &           & \underline{98.51}       & 75.5                               &  & 80                                 & 65.0                              \\
ASPanFormer~\cite{chen2022aspanformer}                               & 24.87                              & 16.2                                    &           & 92.54                                    & 70.4                               &  & 73.33                              & 62.8                              \\
MCNet~\cite{zhu2024mcnet}                                            & 32.89                              & 20.9                                    &           & 92.54                                    & 69.3                               &  & 60                                 & 48.6                              \\
GeoFormer~\cite{liu2023geometrized}                                  & 36.10 & 24.1       &           & \underline{98.51}       & 75.6                               &  & \underline{93.33} & 71.4                              \\ 
RetinaRegNet~\cite{sivaraman2024retinaregnetzeroshotapproachretinal} & 31.28                              & 20.3                                    &           & \textbf{99.25}          & 77.9 &  & \textbf{100}      & 86.8 \\ \hline
\textsc{ART} w/o CAL (ours)  & \underline{51.87}   & \underline{37.2} & \textbf{} & \textbf{99.25} & \underline{78.2}     &  & \textbf{100}      & \underline{92.3}    \\
\textsc{ART} w/ CAL (ours)  & \textbf{64.71}   & \textbf{40.1} &  & \textbf{99.25} & \textbf{78.5}     &  & \textbf{100}      & \textbf{92.5}    \\ \hline
\end{tabular}}
\scriptsize{
The bold and underline values denote the best and second best results, respectively.}
\end{table*}

\subsection{Evaluation on Retinal Categories}\label{subsec:retinal}

\paragraph{Baselines for Comparison}
We compare \textsc{ART} with SuperPoint~\cite{detone2018superpoint}, GLAMpoints~\cite{truong2019glampoints}, ISTN~\cite{lee2019istn}, NCNet~\cite{rocco2020ncnet}, SuperGlue~\cite{sarlin2020superglue}, REMPE~\cite{hernandez2020rempe}, DLKFM~\cite{zhao2021deep}, LoFTR~\cite{sun2021loftr}, IHN~\cite{cao2022iterative}, SuperRetina~\cite{liu2022semi}, ASPanFormer~\cite{chen2022aspanformer}, MCNet~\cite{zhu2024mcnet}, GeoFormer~\cite{liu2023geometrized}, and RetinaRegNet~\cite{sivaraman2024retinaregnetzeroshotapproachretinal}.

\vspace{-10pt}
\paragraph{Evaluation Metrics}
To evaluate alignment performance, we use the CEM approach~\cite{charles2003dual} to calculate the median error (MEE) and maximum error (MAE). 
The results are categorized as follows: 
i) \textit{Acceptable} (MAE \(<\) $50$ and MEE \(<\) $20$),
ii) \textit{Inaccurate} (others).
We also calculated the Area Under Curve (AUC) score~\cite{hernandezmatas2017fire}, with mean AUC (mAUC).

\vspace{-10pt}
\paragraph{Discussion}
We randomly split the KBSMC dataset into 3,370 training and 374 test pairs and trained the model in a FS manner.
For the FIRE~\cite{hernandezmatas2017fire} and FLORI21~\cite{Li2021Flori} datasets, \textsc{ART} was trained in a SS manner using SFIs from KBSMC and FIRE, as well as UWFIs from KBSMC and FLORI21, with warped pairs synthesized via random transformations.

The results in Tab.~\ref{tab:retinal} clearly demonstrate the superiority of the proposed \textsc{ART} method across multiple retinal datasets.
Compared to state-of-the-art methods, \textsc{ART} consistently achieves the highest \textit{Acceptable} rate and mAUC, confirming its effectiveness in retinal image alignment.


On the challenging KBSMC dataset, \textsc{ART} achieves an \textit{Acceptable} rate of 64.71\% and an mAUC of 40.1, outperforming GeoFormer~\cite{liu2023geometrized}.
These results highlight the capability of \textsc{ART} to handle complex retinal image transformations.
On the FIRE and FLORI21 datasets, \textsc{ART} achieves near-perfect \textit{Acceptable} rates of 99.25\% and 100\%, and state-of-the-art mAUCs of 78.5 and 92.5, respectively.

Although methods such as DLKFM~\cite{zhao2021deep} and MCNet~\cite{zhu2024mcnet} adopt iterative point refinement, their reliance on only four points for homography estimation limits their performance, especially on complex, high-resolution pairs like SFI-UWFI. 
In contrast, \textsc{ART} uses a more expressive transformation model, achieving accurate and reliable alignment in challenging scenarios.

Fig.~\ref{fig:FIG_RETINAL} presents qualitative results, further demonstrating the effectiveness of \textsc{ART} compared to state-of-the-art methods such as GeoFormer~\cite{liu2023geometrized} and RetinaRegNet~\cite{sivaraman2024retinaregnetzeroshotapproachretinal}.

\clearpage
\twocolumn




\begin{figure}[t!]
    \centering
    \begin{tikzpicture}
        \node (img1) at (0,0) {\includegraphics[width=0.95\columnwidth]{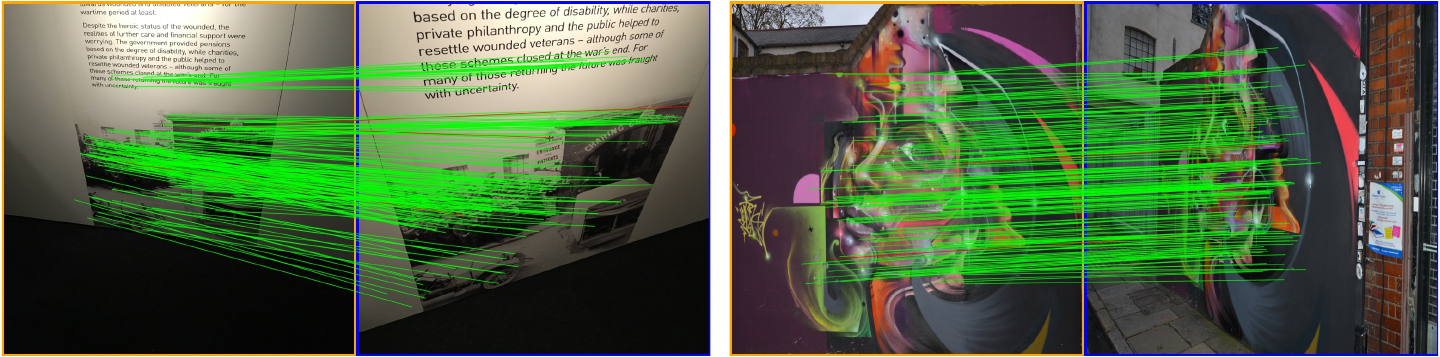}};
        \node (img2) at (0,-2) {\includegraphics[width=0.95\columnwidth]{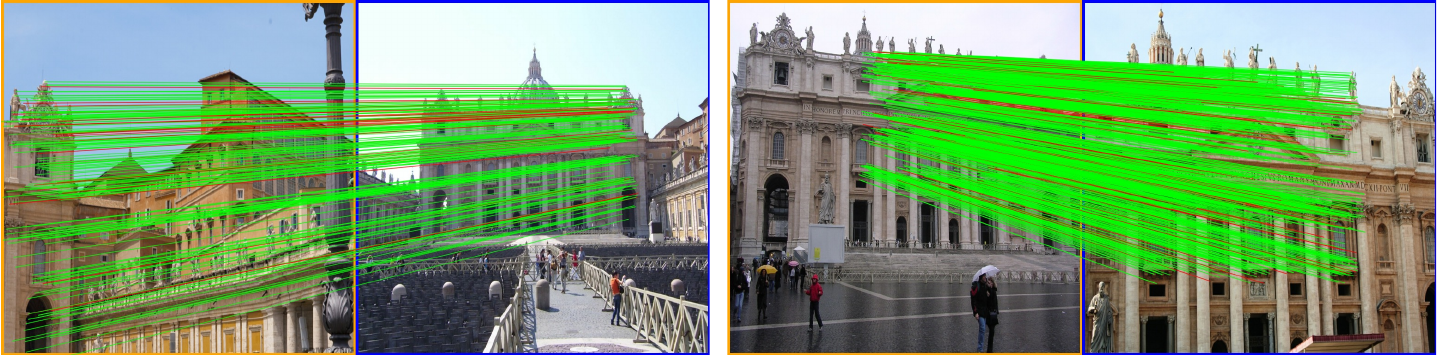}};
        \node (img3) at (0,-4) {\includegraphics[width=0.95\columnwidth]{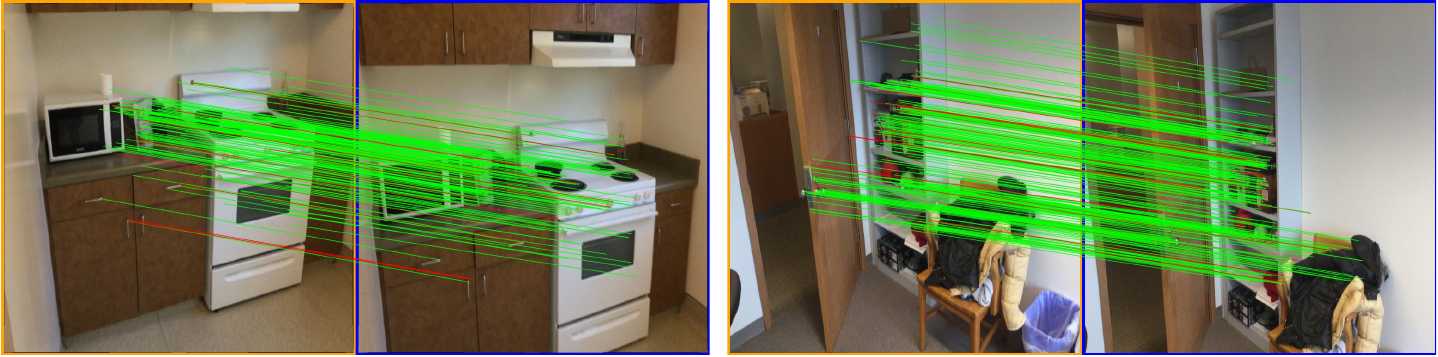}};

        \node[rotate=90, anchor=center] at (-4.2, 0) {\tiny HPatches~\cite{balntas2017hpatches}};
        \node[rotate=90, anchor=center] at (-4.2, -2) {\tiny MegaDepth-1500~\cite{MegaDepthLi18}};
        \node[rotate=90, anchor=center] at (-4.2, -4) {\tiny ScanNet-1500~\cite{dai2017scannet}};
    \end{tikzpicture}

    \caption{
        \textbf{Qualitative Evaluation on Scene-HR Datasets.} To evaluate our method under diverse conditions, we visualize correspondences on the Scene-HR datasets, including HPatches~\cite{balntas2017hpatches}, MegaDepth-1500~\cite{MegaDepthLi18}, and ScanNet-1500~\cite{dai2017scannet}. Correct and incorrect matches are shown in green and red, respectively.
    }
    \label{fig:FIG_HR}
\end{figure}


\begin{table}[h]
{
\caption {\textbf{Quantitative Evaluation on Scene-HR Datasets.}}
\label{tab:scenery_hr}
\resizebox{1.0\columnwidth}{!}{
{\LARGE
\begin{tabular}{clllcccc}
\hline
\multicolumn{3}{c}{\multirow{2}{*}{\textbf{Methods}}}     &  & \multicolumn{3}{c}{\textbf{mAUC}{$\uparrow$}}       &  \\ \cline{5-7}
\multicolumn{3}{c}{}                              
&  & \textbf{HPatches} & \textbf{MegaDepth-1500} & \textbf{ScanNet-1500} &  \\ \hline
\multicolumn{2}{l}{LoFTR~\cite{sun2021loftr}}  
&  &  &    75.4    &   67.7         &     40.7      &    \\
\multicolumn{2}{l}{LightGlue~\cite{lindenberger2023lightglue}}  
&  &  &    77.5    &   72.3         &     48.2      &    \\
\multicolumn{2}{l}{MatchFormer~\cite{wang2022matchformer}}        
&  &  &    78.1    &  68.2          &     43.2      &     \\ 
\multicolumn{2}{l}{RoMa~\cite{edstedt2024roma}}        
&  &  &    \underline{78.4}    &  \textbf{75.2}          &     \textbf{52.0}      &     \\ \hline
\multicolumn{2}{l}{\textsc{ART} w/o CAL (ours)}                
&  &  &    75.3    &  68.8          &     45.8      &   \\
\multicolumn{2}{l}{\textsc{ART} w/ CAL (ours)}                
&  &  &    \textbf{78.6}    &  \underline{74.9}          &     \underline{51.1}      &   \\
\hline
\end{tabular}
}}}
\scriptsize{
The bold and underline values denote the best and second best results, respectively.
}
\end{table}

\subsection{Evaluation on Scene Categories}\label{subsec:scenery}

\paragraph{Baselines for Comparison}
We compare \textsc{ART} with LoFTR~\cite{sun2021loftr}, LightGlue~\cite{lindenberger2023lightglue}, MatchFormer~\cite{wang2022matchformer}, and RoMa~\cite{edstedt2024roma} for HR datasets and DLKFM~\cite{zhao2021deep}, IHN~\cite{cao2022iterative}, and MCNet~\cite{zhu2024mcnet} for LR datasets.

\vspace{-10pt}
\paragraph{Evaluation Metrics}
For 2D geometric transformation datasets, we follow prior works~\cite{zhao2021deep,sun2021loftr} and report the average corner error (ACE) on GoogleEarth~\cite{zhao2021deep}, GoogleMap~\cite{zhao2021deep}, and MSCOCO~\cite{lin2015microsoftcoco}, and the mAUC of ACE on HPatches~\cite{balntas2017hpatches} at thresholds of $3$, $5$, and $10$ pixels.
For two-view transformations (MegaDepth-1500~\cite{MegaDepthLi18}, ScanNet-1500~\cite{dai2017scannet}), we follow prior works~\cite{sun2021loftr,edstedt2024roma} and report the mAUC of pose error at $5^\circ$, $10^\circ$, and $20^\circ$, defined as the maximum angular deviation in rotation and translation.



\begin{figure}[t!]
    \centering

    \begin{minipage}[b]{0.326\linewidth}
        \centering
        \includegraphics[width=\linewidth]{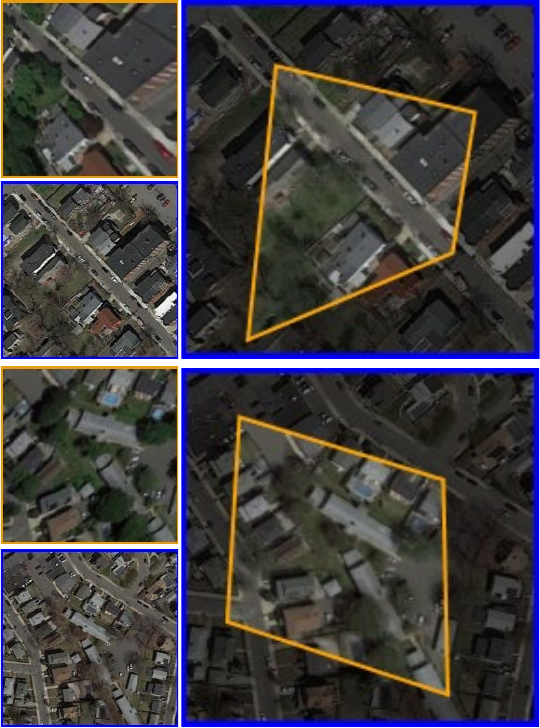}
        \scriptsize GoogleEarth~\cite{zhao2021deep}
    \end{minipage}
    \hfill
    \begin{minipage}[b]{0.326\linewidth}
        \centering
        \includegraphics[width=\linewidth]{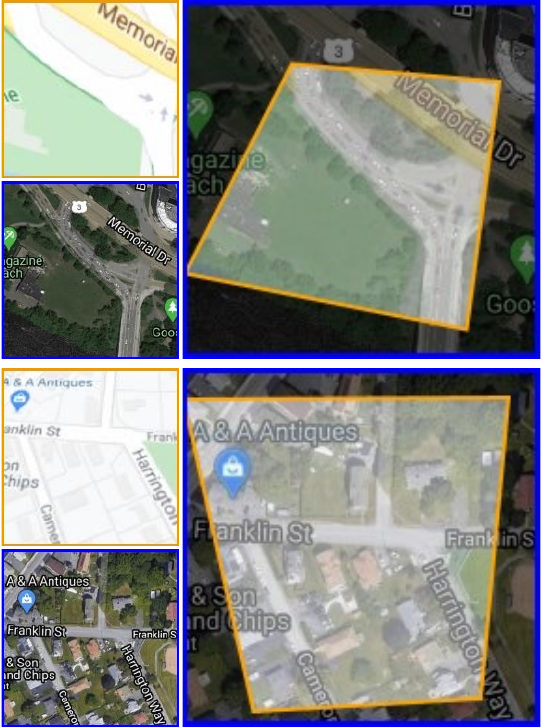}
        \scriptsize GoogleMap~\cite{zhao2021deep}
    \end{minipage}
    \hfill
    \begin{minipage}[b]{0.326\linewidth}
        \centering
        \includegraphics[width=\linewidth]{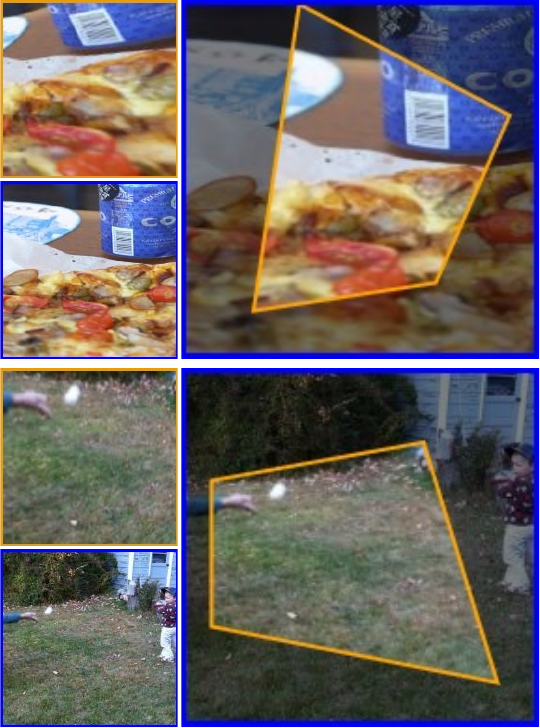}
        \scriptsize MSCOCO~\cite{lin2015microsoftcoco}
    \end{minipage}

    \caption{
        \textbf{Qualitative Evaluation on Scene-LR Datasets.}
        On the GoogleEarth~\cite{zhao2021deep}, GoogleMap~\cite{zhao2021deep}, and MSCOCO~\cite{lin2015microsoftcoco} datasets, \textsc{ART} successfully finds the correct transformation between input image pairs, even with sparse features from low resolution, domain gaps, and scale differences.
    }
    \label{fig:FIG_LR}
\end{figure}

\begin{table}[h!]
{
\caption {\textbf{Quantitative Evaluation on Scene-LR Datasets.}}
\label{tab:scenery_lr}
\resizebox{1.0\columnwidth}{!}{
{\LARGE
\begin{tabular}{clllcccc}
\hline
\multicolumn{3}{c}{\multirow{2}{*}{\textbf{Methods}}}     &  & \multicolumn{3}{c}{\textbf{ACE}{$\downarrow$}}       &  \\ \cline{5-7}
\multicolumn{3}{c}{}                              
&  & \textbf{GoogleEarth} & \textbf{GoogleMap} & \textbf{MSCOCO} &  \\ \hline
\multicolumn{2}{l}{DLKFM~\cite{zhao2021deep}}  
&  &  &    3.88    &   4.41         &     0.55      &    \\
\multicolumn{2}{l}{IHN~\cite{cao2022iterative}}  
&  &  &    1.60    &   0.92         &     0.19      &    \\
\multicolumn{2}{l}{MCNet~\cite{zhu2024mcnet}}        
&  &  &    \underline{0.60}    &  \underline{0.23}          &     \textbf{0.03}      &     \\ \hline
\multicolumn{2}{l}{\textsc{ART} w/o CAL (ours)}                
&  &  &    0.65    &  0.96          &     \underline{0.05}      &   \\
\multicolumn{2}{l}{\textsc{ART} w/ CAL (ours)}                
&  &  &    \textbf{0.17}    &  \textbf{0.19}          &     \textbf{0.03}      &   \\
\hline
\end{tabular}
}}}
\scriptsize{
The bold and underline values denote the best and second best results, respectively.
}
\end{table}

\vspace{-10pt}
\paragraph{Discussion}
For the HR datasets, we pretrained the \textsc{ART} using images from the Oxford-Paris datasets~\cite{4270197,4587635} and finetuned it in a SS manner on the HR datasets.
For the LR datasets, we trained the model in a FS manner.

Tab.~\ref{tab:scenery_hr} and~\ref{tab:scenery_lr} present comparative quantitative evaluations, demonstrating the effectiveness of the proposed \textsc{ART} across HR and LR datasets with varying characteristics.

For the HR datasets, \textsc{ART} achieves state-of-the-art performance on HPatches~\cite{balntas2017hpatches}, which features mostly planar surfaces with rich structural details and minimal domain shifts. 
ART also achieves performance comparable to state-of-the-art on estimating two-view 3D geometric transformations from the MegaDepth-1500~\cite{MegaDepthLi18} and ScanNet-1500~\cite{dai2017scannet} datasets.
We believe the lack of improvement compared to RoMa~\cite{edstedt2024roma} is due to challenges like limited overlap, repetitive structures, and severe degradations that are hard to fully simulate despite extensive training augmentations (Sec.~\ref{subsec:implementation}).
Fig.~\ref{fig:FIG_HR} shows challenging cases where \textsc{ART} accurately estimates correspondences across datasets.

In contrast to HR datasets, LR datasets introduce additional challenges including significant feature loss and resolution discrepancies. 
GoogleMap~\cite{zhao2021deep} images also exhibit domain shift, adding to the alignment difficulty.
As shown in Tab.~\ref{tab:scenery_lr}, traditional iterative deep homography estimation methods~\cite{zhao2021deep,cao2022iterative} struggle under these conditions.
\textsc{ART} achieves the lowest ACE across all datasets, demonstrating superior adaptability to low-resolution and cross-domain alignment tasks.
This suggests that our method’s contextual feature refinement contributes to robust alignment, even in scenarios with large-scale variations and domain mismatches.
Fig.~\ref{fig:FIG_LR} further highlights these capabilities, depicting cases where \textsc{ART} successfully estimates transformations for LR datasets.


\vspace{-12pt}
\subsection{Understanding ART}\label{subsec:ablation}
Here, we present ablation studies to gain a deeper understanding of the key components that constitute \textsc{ART}. 


\begin{figure}[t]
    \centering

    \begin{minipage}{\columnwidth}
        \centering
        \includegraphics[width=\linewidth]{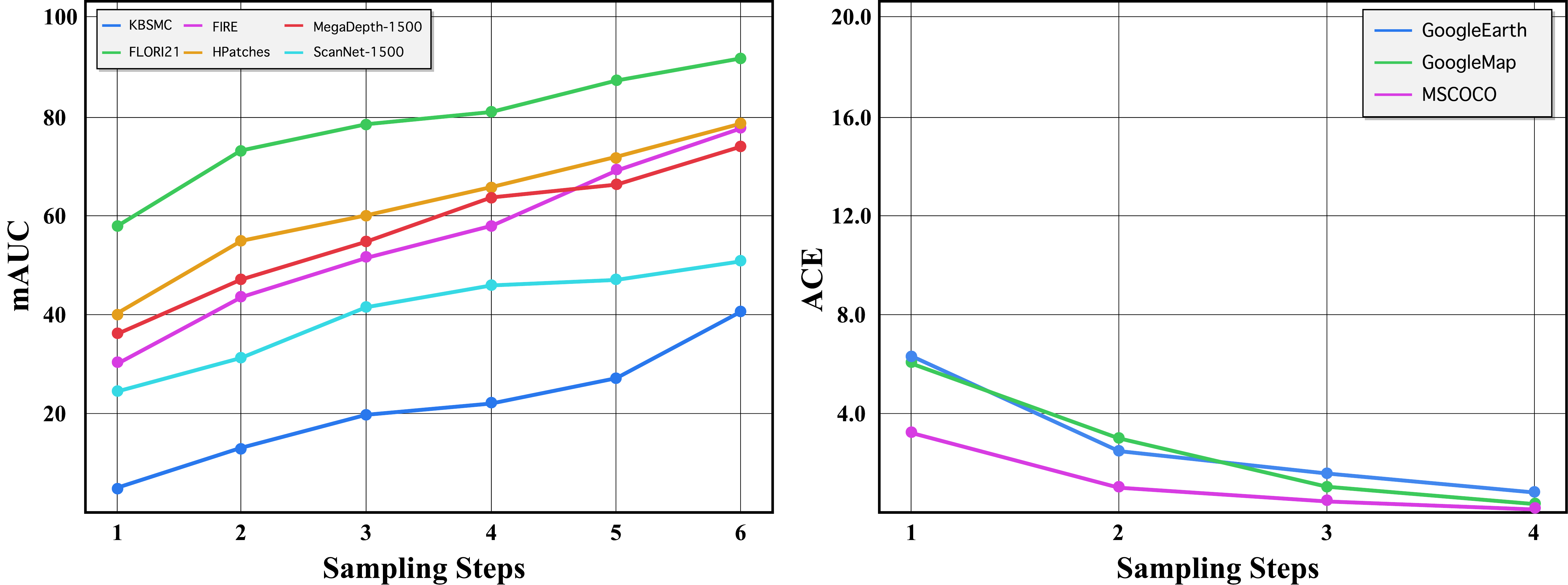}
        
        {\fontsize{6.5pt}{7.5pt}\selectfont (a) Performance change with standard initialization}
    \end{minipage}
    \vspace{0.8em}

    \begin{minipage}{\columnwidth}
        \centering
        \includegraphics[width=\linewidth]{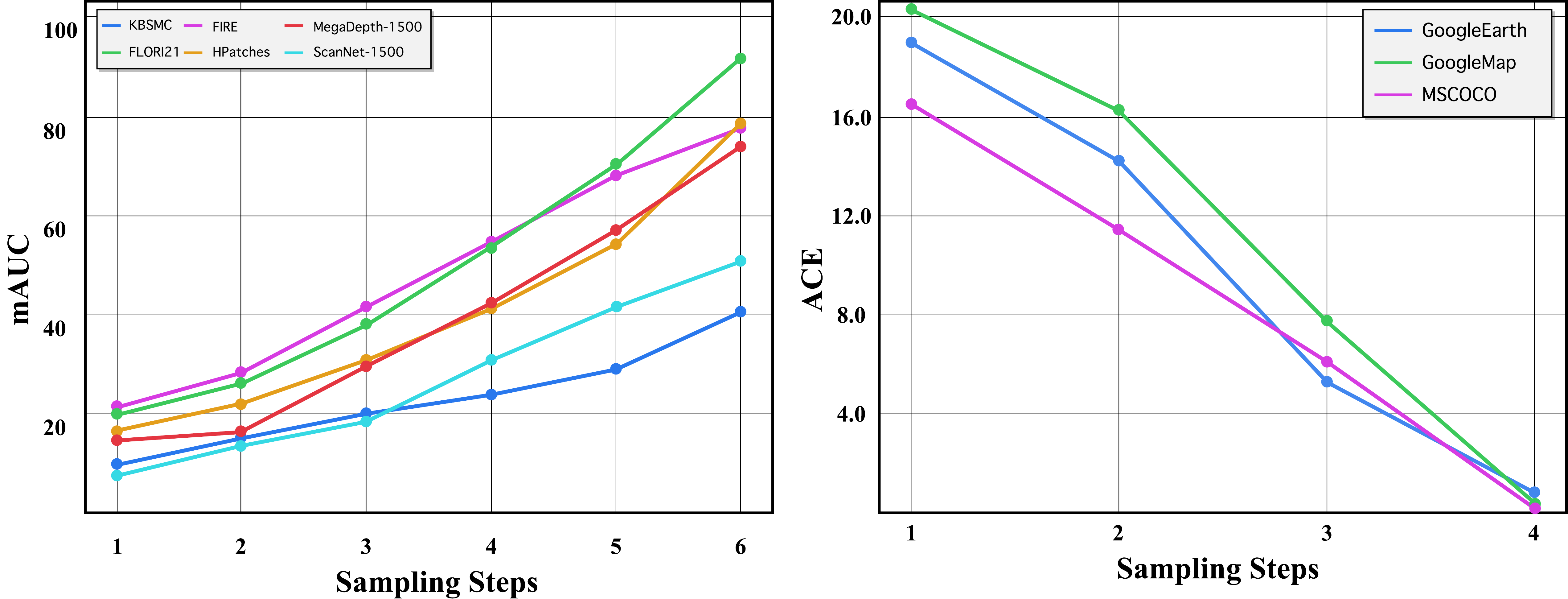}
        {\fontsize{6.5pt}{7.5pt}\selectfont (b) Performance change with varying initialization}
    \end{minipage}

    \caption{
        \textbf{Ablation Study on Sampling.} \textsc{ART} performance varies with (a) the number of sampling steps and (b) different initialization strategies, across HR (left) and LR (right) datasets.
    }
    \label{fig:FIG_ABL}
\end{figure}

\vspace{-10pt}
\paragraph{Sampling Efficiency}
The aforementioned number of iteration steps, $6$ for HR images and $4$ for LR images, can be reduced to improve efficiency during inference. 

For instance, starting with the initialized transform field parameters \(\mathcal{D}_{\mathcal{M}}^0 = \mathbf{1}_{12 \times 12 \times 2}\) and \(\mathcal{D}_{\mathcal{A}}^0 = \mathbf{0}_{12 \times 12 \times 2}\), only a few model inference steps are needed, followed by upsampling to achieve the full resolution of the transform field parameters \( \mathcal{D}_{\mathcal{M}}^K \) and \( \mathcal{D}_{\mathcal{A}}^K \).
The results in Fig.~\ref{fig:FIG_ABL}~(a) show that the network estimates the transform field parameters early in the autoregressive process for both HR and LR datasets without requiring the full number of refinement steps.
This suggests that, compared to other iterative refinement-based approaches~\cite{cao2022iterative,cao2023recurrent,zhu2024mcnet}, our autoregressive process is not only more effective but also adaptable to accelerated sampling.

Alternatively, increasing the spatial size of the initialized transform field parameters \( \mathcal{D}_{\mathcal{M}}^0 \) and \( \mathcal{D}_{\mathcal{A}}^0 \) can also allow \textsc{ART} to estimate the final parameters \( \mathcal{D}_{\mathcal{M}}^K \) and \( \mathcal{D}_{\mathcal{A}}^K \) at full resolution with fewer sampling steps.
In Fig.~\ref{fig:FIG_ABL}~(b), the transform field parameters are initialized at resolutions of \(\mathcal{D}_{\mathcal{M}}^0 = \mathbf{1}_{H / (2^k) \times W / (2^k) \times 2}\) and \(\mathcal{D}_{\mathcal{A}}^0 = \mathbf{0}_{H / (2^k) \times W / (2^k) \times 2}\),
where \(H\) and \(W\) denote the spatial resolution of the input image. 
It is evident that this limits the model’s capacity for wide-range coarse estimation as well as fine-grained local refinements, resulting in poor performance.

\paragraph{Importance of CAL}
Comparative results with and without the use of CAL are reported in Tab.~\ref{tab:retinal}, \ref{tab:scenery_hr}, and \ref{tab:scenery_lr} across all datasets.
Without CAL, \( \tilde{\mathcal{F}}_{s \rightarrow d}^k \) is computed as \(\texttt{Concat}[\mathcal{F}_s^k, \mathcal{F}_d^k]\), lacking the attention necessary for refining the transform field parameters.  

For image pairs in datasets such as FIRE~\cite{hernandezmatas2017fire}, FLORI21~\cite{Li2021Flori}, or HPatches~\cite{balntas2017hpatches}, which contain rich details and abundant features, CAL offers only marginal improvement.
In contrast, for LR datasets with sparse features, such as GoogleEarth~\cite{zhao2021deep}, MSCOCO~\cite{lin2015microsoftcoco}, and GoogleMap~\cite{zhao2021deep}, CAL significantly improves performance.
In the challenging KBSMC dataset, characterized by ambiguous features, large scale variation, and a significant domain gap, CAL notably improves the \textit{Acceptable} rate.

\vspace{-10pt}
\paragraph{Computational Cost}
\textsc{ART} offers markedly better runtime efficiency compared to coarse-to-fine baselines such as LoFTR~\cite{sun2021loftr} (1.101s) and GeoFormer~\cite{liu2023geometrized} (1.150s), due to its lightweight multi-scale design.
On an NVIDIA A100 GPU, \textsc{ART} runs in 0.16s using 261MB of memory.

\vspace{-10pt}
\paragraph{Transform Field Representation}
We empirically observed that predicting only additive parameters for translation degrades performance due to their wide dynamic range.  
Introducing multiplicative parameters for scaling stabilizes model optimization by normalizing spatial displacements, reducing translation variance, and ultimately enhancing performance.
In contrast, directly predicting full affine or projective transform parameters often resulted in non-convergence due to the model’s excessive complexity.

\vspace{-10pt}
\paragraph{Limitations}
In this paper, we have considered only two types of datasets: HR and LR, with spatial resolutions of \(768 \times 768\) and \(192 \times 192\), respectively. 
This setting inherently constrains the input image size, and the required resizing can directly degrade the network’s performance. 
For example, when estimating correspondences between larger images, the predefined $6$ sampling steps may be insufficient to determine the transformation parameters accurately. 
Future research should explore adaptive sampling strategies that adjust dynamically to the input resolution.

\section{Conclusion}
\label{sec:conclusion}

\textsc{ART} tackles the challenging problem of image alignment, where existing methods struggle due to homogeneous textures, large scale differences, and weak feature regions. \textsc{ART} effectively mitigates issues related to poor initialization and scale dependency, achieving precise alignment even in difficult scenarios. Through extensive evaluations across diverse datasets, we demonstrated that \textsc{ART} significantly outperforms existing feature-based, intensity-based, and iterative refinement-based approaches. We believe that \textsc{ART} provides a strong foundation for future research in image alignment, particularly for applications such as medical imaging, remote sensing, and scene analysis.


\vspace{-5pt}
\paragraph{Acknowledgements}
\label{sec:acknowledgements}
\small{This work was supported in part by the IITP grants [No.2021-0-01343, Artificial Intelligence Graduate School Program (Seoul National University), No. 2021-0-02068, No.2023-0-00156, and No.RS-2025-02219317], the NRF grant (RS-2025-00521972, AI Star Fellowship (Kookmin University)), and the Industrial Technology Alchemist Project [No. RS-2024-00432410] funded by MOTIE, Korea.}
{
    \small
    \bibliographystyle{unsrt}
    \bibliography{main}
}

\end{document}